\documentclass[lettersize,journal]{IEEEtran}
\usepackage{amsmath,amsfonts}
\usepackage{algorithmic}
\usepackage{algorithm}
\usepackage{array}
\usepackage{textcomp}
\usepackage{stfloats}
\usepackage{url}
\usepackage{array,multirow,graphicx}
\usepackage{subcaption}
\usepackage{float}
\usepackage{gensymb}
\usepackage{caption}
\usepackage{placeins}
\usepackage{multirow}
\usepackage{flushend}
\usepackage{verbatim}
\usepackage{hyperref}
\usepackage{graphicx}
\usepackage{cite}
\graphicspath{{./Fig/}}	
\usepackage[english]{babel}
\DeclareMathOperator{\arctg}{arctg}
\DeclareMathOperator{\ch}{ch}
\DeclareMathOperator{\sh}{sh}
\DeclareMathOperator{\tg}{tg}
\DeclareUnicodeCharacter{2061}{}

\begin{document}

\title{\LARGE \bf Dexterous Manipulation of Deformable Objects \\ via Pneumatic Gripping: Lifting by One End}

\author{Roman Mykhailyshyn,~\IEEEmembership{Member,~IEEE,} Jonathan Lee, Mykhailo Mykhailyshyn, \\ Kensuke Harada,~\IEEEmembership{Fellow,~IEEE,} and Ann Majewicz Fey,~\IEEEmembership{Member,~IEEE}

\thanks{Roman Mykhailyshyn is with the Graduate School of Engineering Science, Osaka University, Osaka 560-8531, Japan, and with transfer to the Industrial Cyber-Physical Systems Research Center at the National Institute of Advanced Industrial Science and Technology (AIST), Japan, and on leave from the Walker Department of Mechanical Engineering, The University of Texas at Austin, Austin, TX 78712, USA. (Corresponding author e-mail: mykhailyshyn.roman.es@osaka-u.ac.jp)}
\thanks{Jonathan Lee is with the Walker Department of Mechanical Engineering, The University of Texas at Austin, Austin, TX 78712, USA. (e-mail: jl0409@utexas.edu)} 
\thanks{Mykhailo Mykhailyshyn is with the Department of Information Science and Mathematical Modeling, Ternopil Ivan Puluj National Technical University, Ternopil, 46001, Ukraine. (e-mail: mms000@ukr.net)}
\thanks{Kensuke Harada is with the Graduate School of Engineering Science, Osaka University, Toyonaka, Osaka 560-8531, Japan. (e-mail: harada@sys.es.osaka-u.ac.jp)}%
\thanks{Ann Majewicz Fey is with the Walker Department of Mechanical Engineering, The University of Texas at Austin, Austin, TX 78712, USA, is also with the Department of Surgery, UT Southwestern Medical Center, Dallas, TX 75390, USA. (e-mail: Ann.MajewiczFey@utexas.edu)}} 



\maketitle

\begin{abstract}
Manipulating deformable objects in robotic cells is often costly and not widely accessible. However, the use of localized pneumatic gripping systems can enhance accessibility. Current methods that use pneumatic grippers to handle deformable objects struggle with effective lifting. This paper introduces a method for the dexterous lifting of textile deformable objects from one edge, utilizing a previously developed gripper designed for flexible and porous materials. By precisely adjusting the orientation and position of the gripper during the lifting process, we were able to significantly reduce necessary gripping force and minimize object vibration caused by airflow. This method was tested and validated on four materials with varying mass, friction, and flexibility. The proposed approach facilitates the lifting of deformable objects from a conveyor or automated line, even when only one edge is accessible for grasping. Future work will involve integrating a vision system to optimize the manipulation of deformable objects with more complex shapes.
\end{abstract}


\section{Introduction}
Expanding the possibility of using robotic systems in all spheres of human activity takes on a new development every year. The new capabilities of computer vision \cite{nicola2024co, coltraro2023robotic, lee2024grasp, longhini2024unfolding, 10417136} and novel gripping devices \cite{papadopoulos2023deformable, 10680377} allow us to look from a different point of view at the previously difficult-to-implement processes of manipulating deformable objects. However, since deformable materials can have various characteristics that are critically different from each other \cite{ahmad2020fibers, 10436531} that makes performing fully automated manipulation still challenging.

Robotic manipulation of deformable objects can be achieved through various methods~\cite{sanchez2018robotic, li2018model, mcconachie2018estimating, arriola2020modeling}, which are significantly influenced by the grasping technique~\cite{koustoumpardis2004review, carbone2012grasping, fantoni2014grasping, mykhailyshyn2022three, borras2020grasping, Mykhailyshyn2022rev}, the type of objects being handled~\cite{spiers2016single}, the operating principles of the gripping device~\cite{10680377, Mykhailyshyn2022gripper, mykhailyshyn2023fem, donaire2020versatile, mykhailyshyn2023optimization, schaler2017electrostatic, wang2021model}, and the strategies for maintaining control over the deformable object~\cite{hanamura2024fabric, zhang2024achieving, Mykhailyshyn2024toward, yoon2022elongatable}.



Textile materials are among the most challenging deformable objects to automate. Key challenges in designing robotic cells for handling textiles include the material's varying flexibility in 3D space \cite{longhini2021textile}, its extreme porosity \cite{swery2016efficient}, and the significant influence of external factors on its behavior \cite{rieber2013influence}, among others. The authors of the article \cite{borras2020grasping} presented the general concept of methods of manipulating textile materials using mechanical grippers.

\begin{figure}[!t]
    \centering
    \vspace{-5mm}
    \includegraphics[width=0.98\linewidth,clip ,trim=35pt 20pt 20pt 0pt]{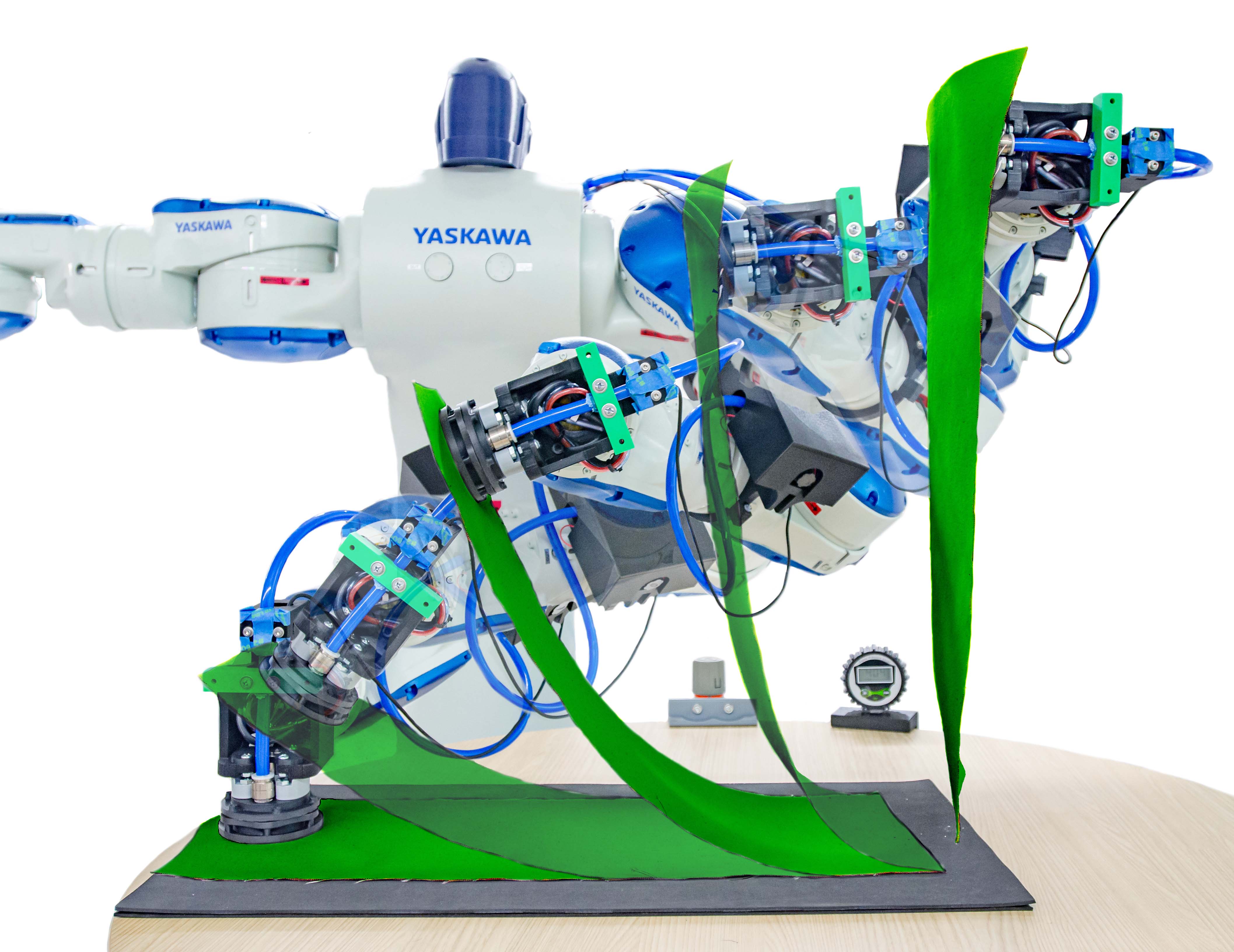}
    \caption{Dexterous lifting of the textile object by one end with pneumatic gripper (Gripping Device for Flexible and Porous Materials~\cite{Mykhailyshyn2022gripper} with 185 kPa supply pressure).}
    \label{fig1}
\end{figure}

Promising research directions for textile manipulation include human-robot interaction \cite{andronas2021model, donner2017cooperative}, advanced sensing techniques \cite{kawaharazuka2019dynamic}, and the use of dual robotic arms \cite{hu2018three, yin2021modeling, miller2012geometric, lv2022dynamic}. However, there are many scenarios where only one robotic arm can be utilized \cite{mira2015study}, often due to spatial constraints or limited access to the textile material, which is a common challenge in manufacturing. Consequently, the development of dexterous single-arm manipulation of deformable objects is a highly promising research topic \cite{bai2016dexterous, andronas2022perception, makris2022deformable, digumarti2021dexterous, puhlmann2022rbo}.

Most current studies focus on using mechanical or hybrid gripping devices (mechanical, electrostatic, soft, etc.) for dexterous manipulation of deformable objects. However, pneumatic grippers are emerging as a promising alternative \cite{Mykhailyshyn2022gripper, mykhailyshyn2022influence, fleischer2016sustainable, reinhart2011flexible}, offering solutions to many challenges in handling textile materials, Robotextile is a prime example. Despite this potential, the effectiveness of pneumatic grippers for manipulating deformable objects in robotic cells remains underexplored and requires further development. This is evident in the limited but growing application of pneumatic gripping systems in textile manufacturing.

\begin{figure}[tb]
\newcommand{\mywidth}{1}
\centering
 	\begin{subfigure}[b]{1\linewidth}
         \centering
         \vspace{2mm}
        \includegraphics[width=\mywidth\linewidth, clip, trim=0pt 0pt 0pt 0pt]{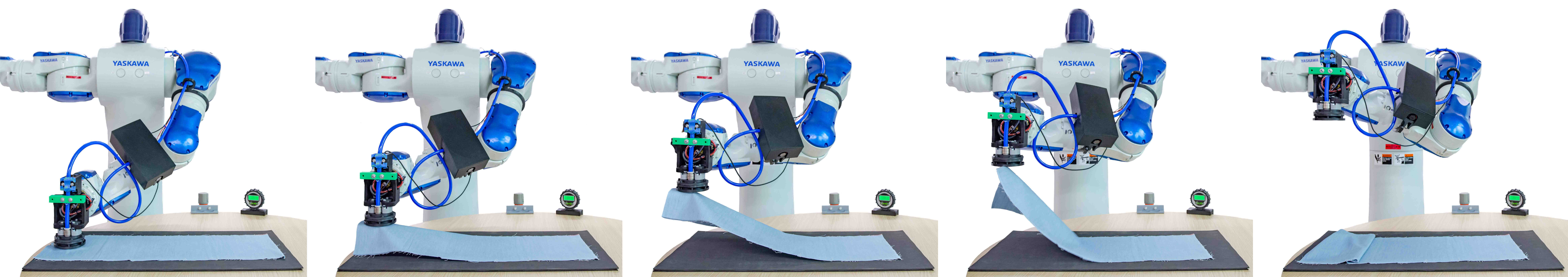}
        \vspace{-5mm}
        \caption{}
        \label{subfig1}
	\end{subfigure}\\
  	\begin{subfigure}[b]{1\linewidth}
         \centering
         \vspace{1mm}
         \includegraphics[width=\mywidth\linewidth, clip, trim=0pt 0pt 0pt 0pt]{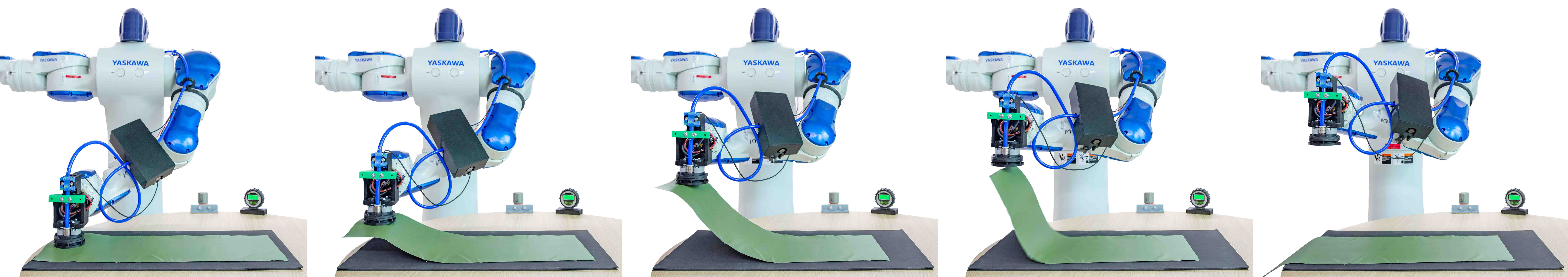}
         \vspace{-5mm}
        \caption{}
        \label{subfig2}
	\end{subfigure}\\
        \vspace{-1mm}
 \caption{Failing classical vertical lifting using a pneumatic gripping device for porous and textile objects: (a) thin denim cotton; (b) 600-denier cordura outdoor canvas waterproof fabric.}
 \label{fig2}
\end{figure}

In our previous work~\cite{Mykhailyshyn2024toward}, we used a gripping device designed for flexible and porous materials~\cite{Mykhailyshyn2022gripper} to explore the challenges of using pneumatic systems to manipulate deformable objects like films and textiles. During a standard vertical lift from one end (Fig.~\ref{fig2}), the object's center of mass shifts, leading to deformation at the gripping edge. This object deformation results in several issues: depressurization of the gripper and reduced lifting force, a minimized contact area with lower friction, and an apparent increase in the object's mass. Regardless of the porosity or mass of deformable objects (Fig.\ref{fig2}), the negative factors outlined will ultimately cause the object to detach from the gripper and fall, despite the pressure supplied to the gripper.

\begin{figure}[!t]
\newcommand{\mywidth}{1}
\centering
        \begin{subfigure}[b]{0.64\linewidth}
         \centering
         \vspace{-4.5mm}         \includegraphics[width=\mywidth\linewidth, clip, trim=0pt 0pt 0pt 0pt]{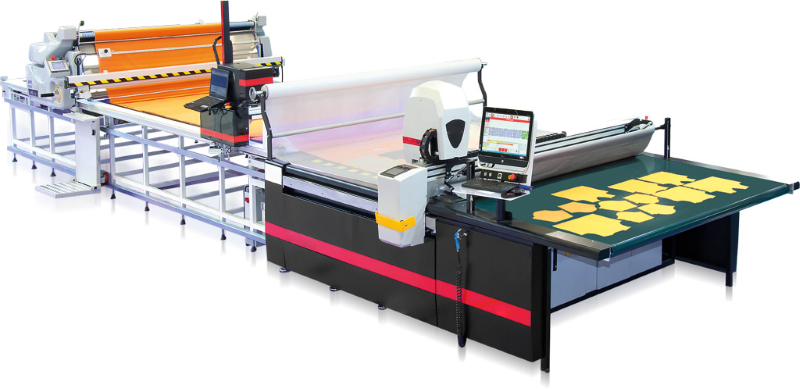}
         \vspace{-5mm}
         \caption{}
        \label{fi1_1}
	\end{subfigure}
            ~
        \begin{subfigure}[b]{0.26\linewidth}
         \centering
         \vspace{-4.5mm} 
         \includegraphics[width=\mywidth\linewidth, clip, trim=0pt 0pt 0pt 0pt]{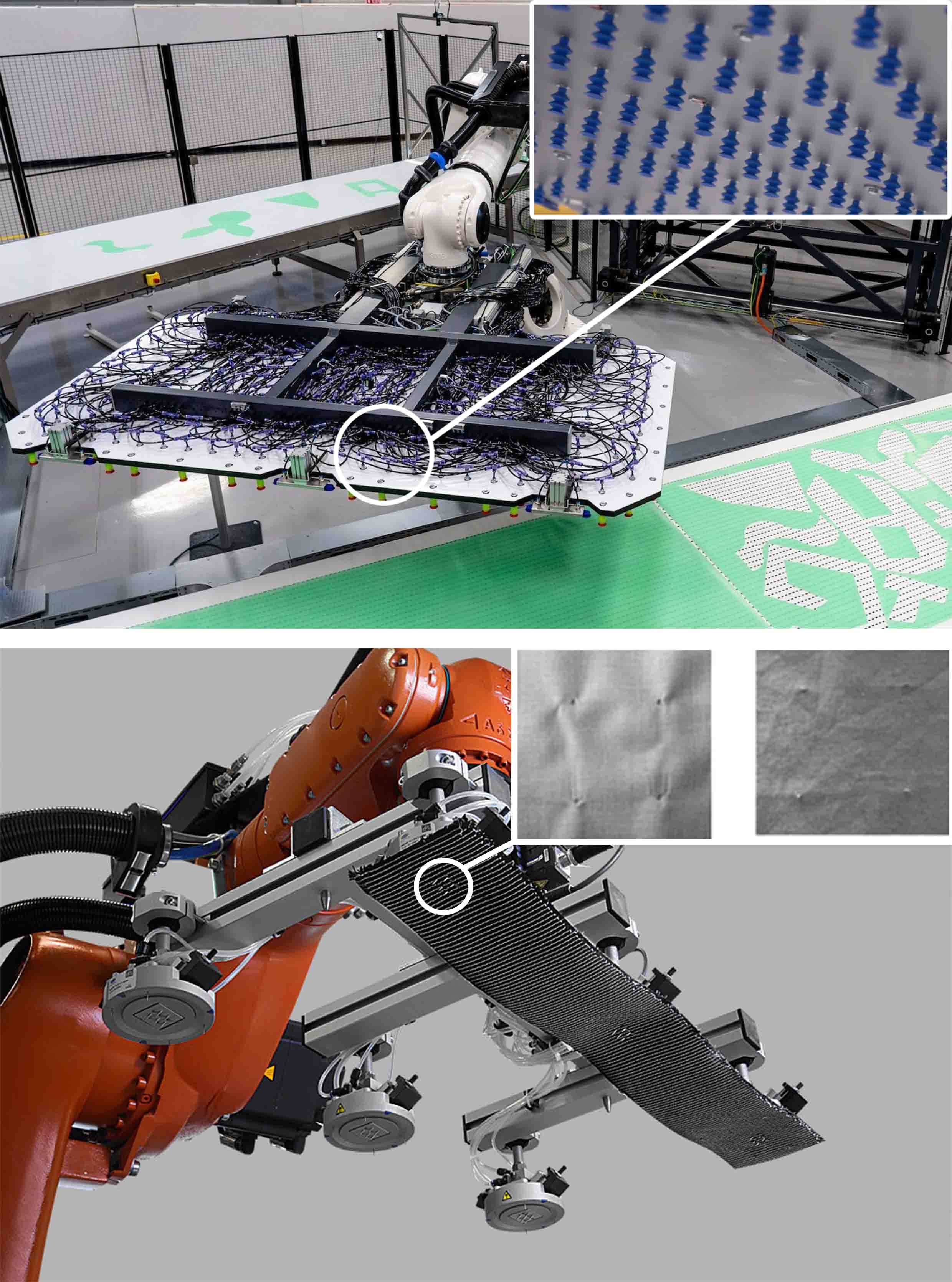}
         \vspace{-5mm}
         \caption{}
        \label{fi1_2}
	\end{subfigure}\\
 
  	\begin{subfigure}[b]{0.88\linewidth}
         \centering         \includegraphics[width=\mywidth\linewidth, clip, trim=0pt 0pt 0pt 0pt]{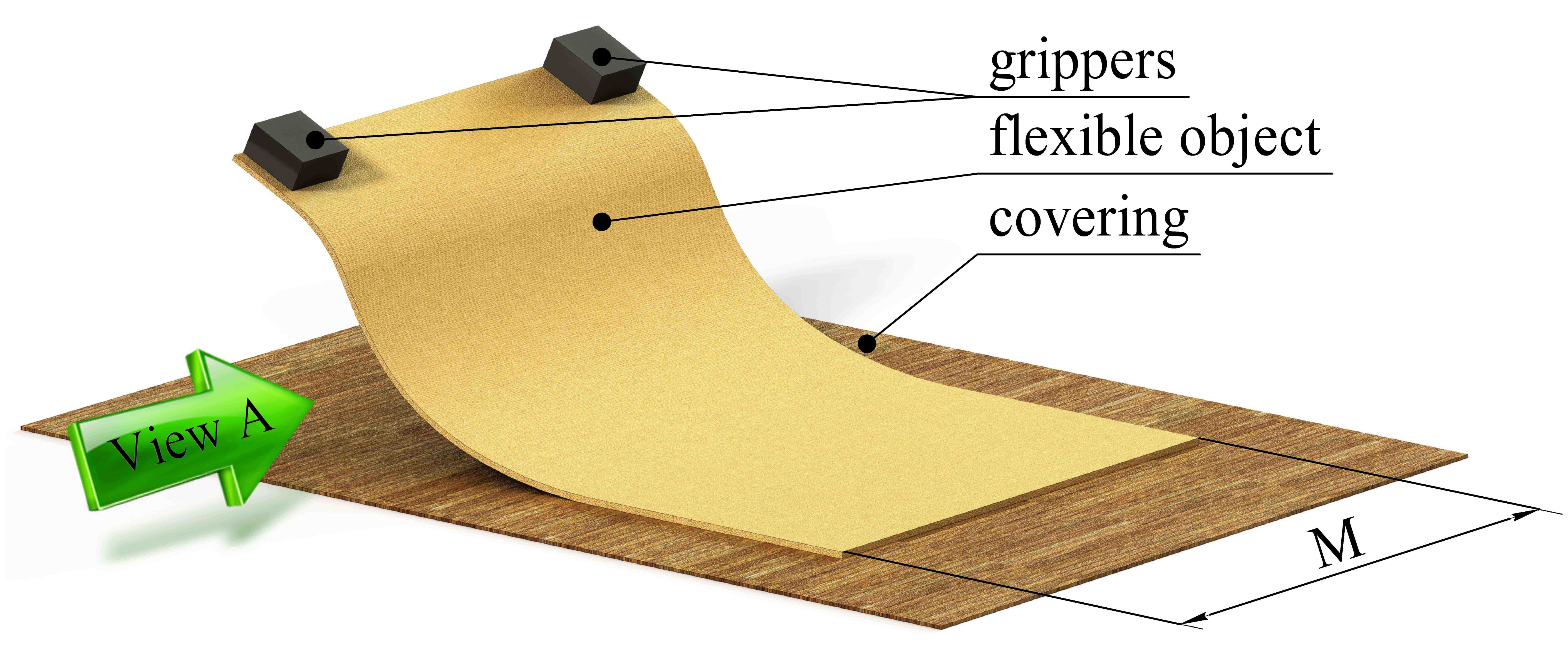}
         \vspace{-7mm}
        \caption{}
        \label{fig1_3}
	\end{subfigure}\\
        \vspace{-1mm}
\caption{Existing and studied methods of grasping/manipulating deformable objects: (a) automated stage of the textile cutting machine (Serkon); (b) commercially available pneumatic grasping system for deformable objects Airborne and Schmalz; (c) manipulation deformable objects using pneumatic grippers with planar contacts and plane.}
\label{fig_1}
\end{figure}

Automating the robotic lifting of deformable objects from one end offers a solution to the challenges of sorting and assembling textiles and other materials after cutting (Fig.\ref{fi1_1}). While mechanical grippers~\cite{yamazaki2021versatile} are commonly used for manipulating deformable objects, they often struggle to separate individual materials or detach objects from conveyor surfaces. Pneumatic grippers also exist for these tasks (Fig.\ref{fi1_2}), but they tend to be either energy-intensive or prone to damaging materials through the use of needles~\cite{ebraheem2021universal}. To address these limitations, the authors in~\cite{Mykhailyshyn2024toward} propose adjusting the orientation of the gripper before movement begins. This approach prevents depressurization and enhances holding force by utilizing friction between the object and the gripper. The authors have successfully applied similar techniques for manipulating flat, rigid objects along both straight~\cite{savkiv2017orientation} and curved trajectories~\cite{savkiv2018modeling}. By taking into account inertia, friction, and frontal resistance forces~\cite{mykhailyshyn2018analysis}, provides opportunities to increase grip stability while minimizing energy consumption~\cite{mykhailyshyn2017experimental} during pick-and-place operations in robotic cells.

In a previous study~\cite{Mykhailyshyn2024toward}, we found that the holding force increases with the increasing angle of orientation of a pneumatic gripper relative to the tension of the deformable object. However, if we consider the entire process of lifting a deformable object, it is obvious that we need to take into account the processes of deformation and interaction of the object with other surfaces, which has not been done before. If we consider that we need to reorient the gripper before starting to lift the deformable object from a certain surface (Fig.~\ref{fig1_3}), we will cause unwanted issues: 
\begin{itemize}
\item Due to the additional stage of reorientation of the grasping system, the time required to pick up the object increases.
\item During reorientation, a large frictional force between the object and the surface (covering) on which it lies begins to act, which causes the object to slide over the gripper, which in turn leads to an increase in the minimum required holding force (by increasing the gripper supply pressure).
\item Since the gripping device~\cite{Mykhailyshyn2022gripper} pushes air to the side, as a result of the reorientation of the gripper, this air collides with the object and causes it to vibrate, which minimizes the contact between the gripper and the object, which in turn leads to an increase in the minimum required holding force (by increasing the gripper supply pressure).
\end{itemize}


In this article, we address the challenges by introducing a novel method for dexterous lifting of deformable objects from one end using pneumatic grippers (Fig.~\ref{fig1}). The proposed approach optimizes the position and orientation of the end effector during the lifting process, reducing airflow interactions with the object, and preventing air resistance forces that could cause detachmentof deformable material. This designed trajectory effectively eliminates friction and tension between the object and its covering, significantly reducing the holding force and required gripper pressure by 19\% to 76\% across the tested materials.

\section{Methodology and Materials}
\begin{figure}[!t]
    \centering
    \vspace{4mm}
    \includegraphics[width=0.84\linewidth, clip, trim=0pt 0pt 0pt 0pt]{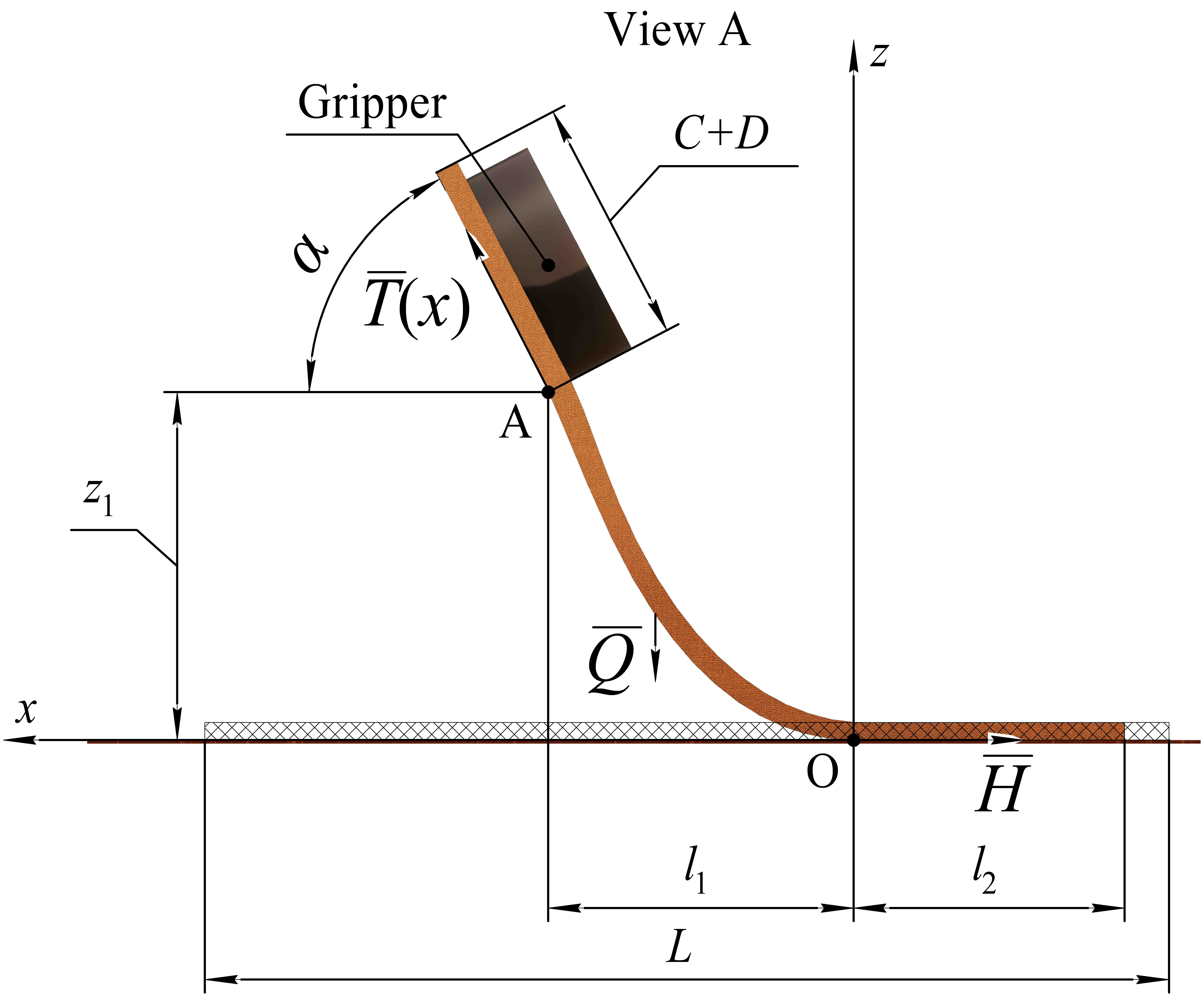}
    \caption{Scheme of the vertical lifting of the object (View A in Fig.~\ref{fig1_3}): $C$~-~distance between the edge of the deformable object to the gripper; $D$~-~diameter of the gripper; $L$~-~length of the deformable object; $z_1$~-~coordinates of point A along the $z$-axis; $l_1$~-~length of the deformed segment AO of the object projected on the $x$-axis; $l_2$~-~length of the object segment lying on the covering; $\alpha$~-~angle of gripper orientation relative to the $x$-axis; $\vec{H}$~-~tension force formed by the friction force of the lying segment of the object ($l_2$); $\vec{Q}$~-~force is equivalent to the mass of the segment of the object AO; $\vec{T}(x)$~-~tension force at point A that equivalent force $\vec{Q}$.}
    \label{fig_2}
\end{figure}

\noindent
Consider a piece of deformable material (such as textile) with a width $M$ and length $L$ lying on a horizontal table/conveyor (Fig.~\ref{fig1_3}). The material is grasped by its edge using multiple grippers. The number of grippers is determined by the width $M$ to prevent sagging between them and ensure sufficient lifting force to hold the material. Thus, the problem of vertically lifting the material with the gripper can be simplified as a two-dimensional problem with a single gripper (Fig.~\ref{fig_2}).

During the upward movement of the gripper, part of the material lying on the table will move in the horizontal direction, since the length of the segment AO is greater than $l_1$. At any moment in time, the section of material AO is a chain line - catenary\footnote{Catenary - is the curve that an idealized hanging chain or cable assumes under its own weight when supported only at its ends.}. If we choose a coordinate system with the origin at the lowest point of the curve and assume that the material deforms as a catenary, then the equation of this curve will look like:
\begin{equation}
\label{deqn_ex1}
z = \frac{H}{q} \bigg( \ch\bigg(\frac{qx}{H}\bigg)-1 \bigg),
\end{equation}

\noindent where $q$ - weight per unit length of material, $H$ - material tension at the lowest point O, ch - hyperbolic function cosh$(x) = (e^x+e^{-x})/2$, and sh - hyperbolic function sinh$(x) = (e^x-e^{-x})/2$. The tension of the material at any point is directed tangentially to the curve. If we consider a smaller part of the material section AO Fig.~\ref{fig_2}, then it is acted upon by the tension force $\vec{H}$ on the end face O, which is directed horizontally, the tension force $\vec{T}$ at point A and the equivalent force of its own weight $\vec{Q}$, which is directed vertically down.

The tension force $\vec{T}$ at any point along the curve, including point A, is directed tangentially. From the equilibrium equation, $T\cos\alpha = H$ is derived by balancing the sum of the force projections on the $x$ axis. The projection of the tension force on the $x$ axis remains constant across all sections and is equal to the tension at the lowest point of the curve. As the gripper lifts the object upward, the inclination angle of the tangent at point A continuously changes, leading to depressurization and a reduction in holding force for pneumatic grippers. Therefore, as shown in the authors' previous work~\cite{Mykhailyshyn2024toward}, the gripper must be oriented tangentially to the edge of the object to maximize holding force. We propose a method where the gripper should be aligned parallel to the force $\vec{T}$ at the extreme point A. The goal, then, is to determine the orientation angle of the tangent at the edge of the material with respect to the $x$ axis for various heights $z_1 = 0...L$. We denote $a=H/q$, and the coordinates of A by $x=l_1$, $z=z_1$, then from the equation (\ref{deqn_ex1}) we find:
\begin{equation}
\label{deqn_ex2}
z_1 = a \bigg( \ch \bigg(\frac{l_1}{a}\bigg)-1 \bigg),
\end{equation}
\noindent or
\begin{equation}
\label{deqn_ex3}
a = \frac{z_1}{\ch \big(\frac{l_1}{a}\big)-1}.
\end{equation}

We found the length of the section AO of the material, which we denote by $L_1$:
\begin{equation}
\label{deqn_ex4}
  \begin{aligned}
L_1 = \int_{O}^{A} \,ds=\int_{O}^{A}\sqrt{1+z\prime^2}dx \,= \\
 =\int_{0}^{l_1}\sqrt{1+\sh ^ 2\bigg(\frac{x}{a}\bigg)} \,dx= a \sh \bigg(\frac{l_1}{a}\bigg).\\
 \end{aligned}
\end{equation}

The length of the part of the material lying on the horizontal plane is $l_2=L-L_1$. The tension force of the material at the lowest point O is equal to the maximum frictional force of the material of length $l_2$. If we denote the coefficient of sliding friction by $k$, then:

\begin{equation}
\label{deqn_ex5}
H = ql_2f=q(L-L_1)k,
\end{equation}
\noindent or
\begin{equation}
\label{deqn_ex6}
\frac{H}{q} = a = (L-L_1)k = Lk-ak \sh \bigg(\frac{l_1}{a}\bigg).
\end{equation}

The equation (\ref{deqn_ex6}) can be written as:
\begin{equation}
\label{deqn_ex7}
a = \frac{Lk}{1 + k \sh \big(\frac{l_1}{a}\big)}.
\end{equation}

As a result, we get a system of two nonlinear equations (\ref{deqn_ex3}) and (\ref{deqn_ex7}) with respect to $a$ and $l_1$. It can be seen that if we subtract another from one equation, we get an equation that includes the ratio $l_1/a$:
\begin{equation}
\label{deqn_ex8}
\frac{\ch \big(\frac{l_1}{a}\big)-1}{1 + k \sh \big(\frac{l_1}{a}\big)} = \frac{z_1}{L k}.
\end{equation}

Solving this equation, we find the ratio $l_1/a$. Substituting this ratio into the right-hand side of the equation (\ref{deqn_ex3}), we find $a$, and therefore $l_1$. The tangent of the tangent angle to the curve at the point $x$ has the value $dz/dx=z^{\prime}(x)= \sh (x/a)$. This means that the tangent of the angle at which we need to orient the gripper at the highest point A will be equal to:

\begin{equation}
\label{deqn_ex9}
\tg \alpha = \sh  \bigg(\frac{l_1}{a}\bigg),
\end{equation}
\noindent or angle at any point of time:
\begin{equation}
\label{deqn_ex10}
\alpha_t =  \arctg  \bigg(\sh \bigg( \frac{l_1}{a}\bigg) \bigg).
\end{equation}

\begin{figure}[!t]
    \centering
    \vspace{4mm}
    \includegraphics[width=0.94\linewidth, clip, trim=0pt 0pt 0pt 0pt]{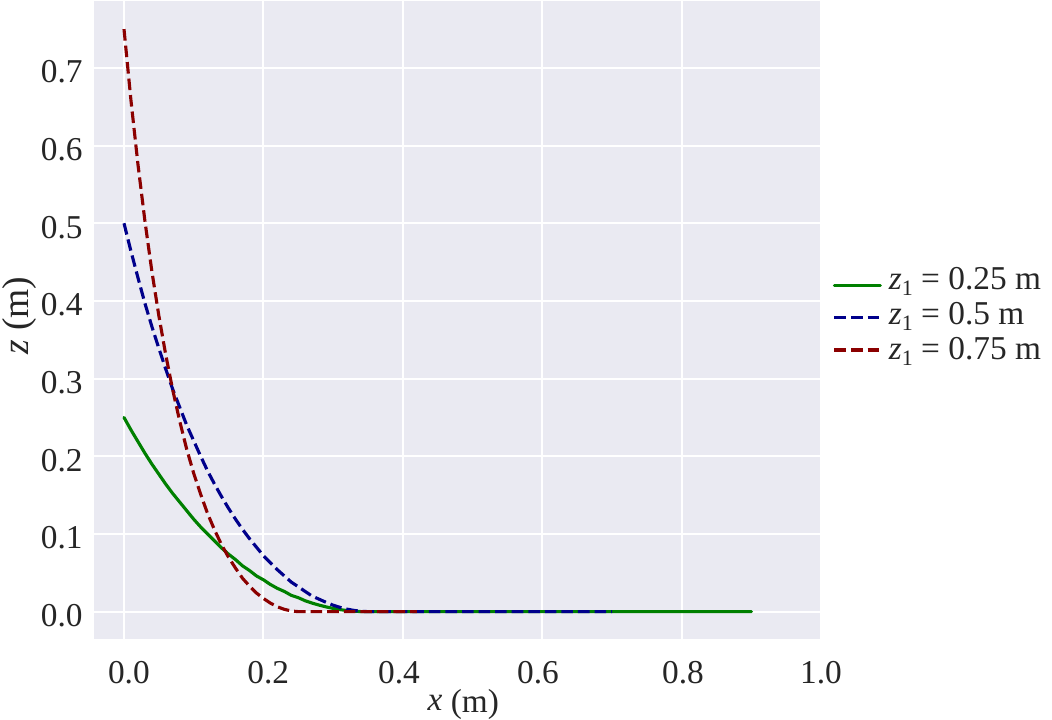}
    \vspace{-2mm}
    \caption{Changing the position and shape of the deformable material as it moves upward ($L = 1$ m, $q = 1$ N/m, and $k = 0.2$)}
    \label{fig_4}
\end{figure}

Using this model, we can determine the position of the deformable material in space during vertical upward movement (Fig.~\ref{fig_4}). As can be seen from Fig.~\ref{fig_4}, the point of the edge of the gripper A when moving to a height of 0.25, 0.5, and 0.75 m leads to the fact that the edge of the fabric moves closer to the axis of vertical movement of the edge of the material. Such movement of the deformable material will lead to its sliding along the horizontal surface. Due to the force of sliding friction, the required holding force will increase, which will lead to additional energy costs for the performance of pick and place operations. 

\begin{figure}[!t]
    \centering
    \vspace{-3mm}
    \includegraphics[width=0.86\linewidth, clip, trim=0pt 0pt 0pt 0pt]{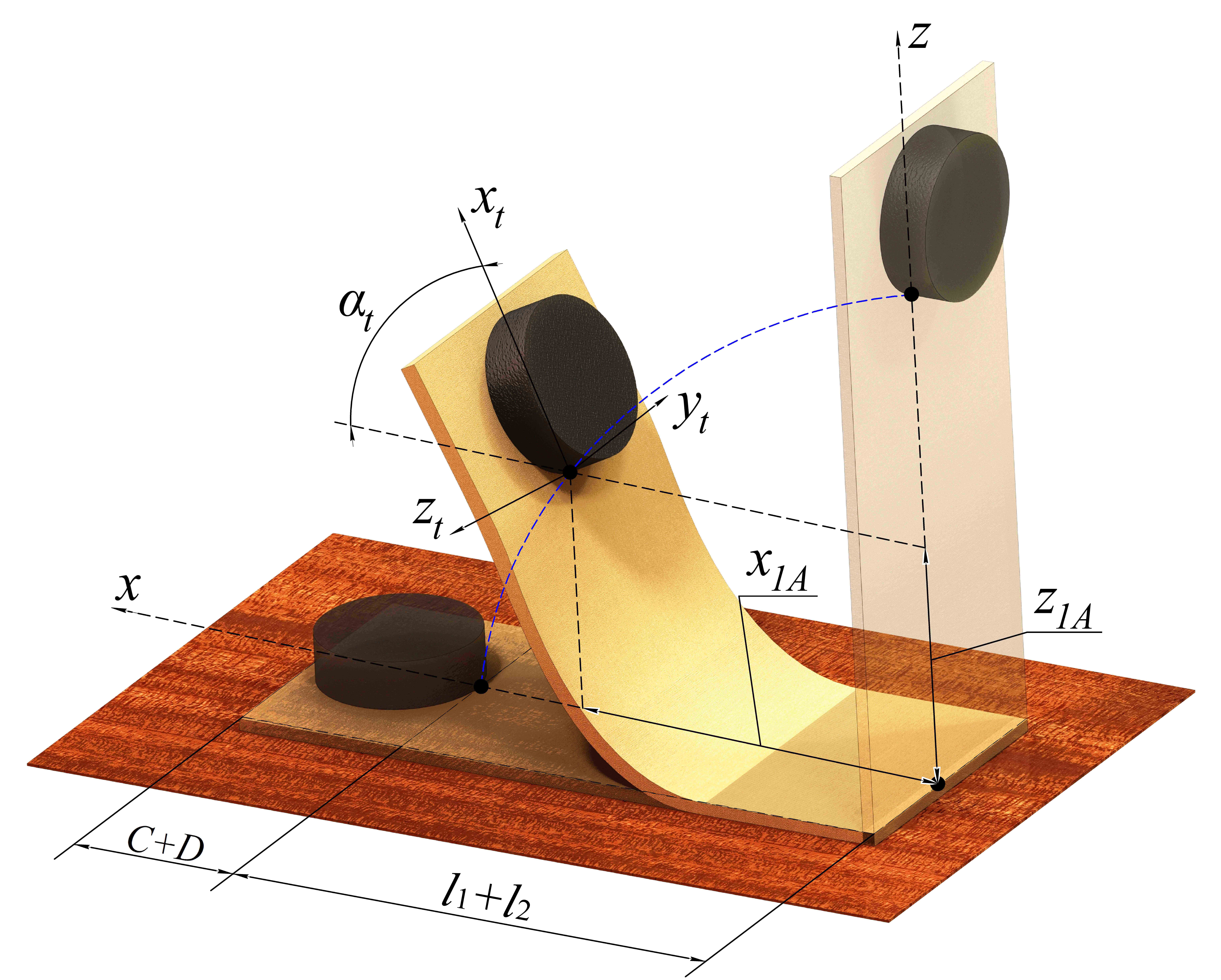}
    \vspace{-2mm}
    \caption{Scheme of dexterous manipulation of deformable objects.}
    \label{fig_5}
\end{figure}

Therefore, the proposed algorithm consists of ensuring a stable reorientation of the gripping device and avoiding sliding of the deformable material on the horizontal plane. To do this, it is suggested to move the end of the hanging material not along a vertical trajectory, but to move it relative to the $x$-axis in the direction of the end of the material by the distance by which the material would have to slip in case vertical movement (Fig.~\ref{fig_5}). The coordinates of the proposed trajectory for the dexterous lifting of the deformable object (point A) are calculated according to the equations (where the origin of the coordinates is taken at the end of the deformable object):

\begin{equation}
\label{deqn_ex11}
x_{1A} = L-\frac{H}{q} \sh \bigg(\frac{l_1 q}{H}\bigg) +l_1,
\end{equation}

\begin{equation}
\label{deqn_ex12}
z_{1A} = \frac{H}{q}  \bigg(\ch \bigg(\frac{l_1 q}{H}\bigg) -1\bigg).
\end{equation}

\subsection{Trajectory planning algorithm}
\noindent Using the above-described methodology for calculating a position and orientation end-effector for manipulating deformable objects with known material parameters such as length ($L$), coefficient of friction ($k$), position, and orientation in a robotic cell, an Algorithm~\ref{alg1} for generating such a trajectory is presented.  

\begin{algorithm}
    \caption{Generate Trajectory}
    \label{alg1}
    \begin{algorithmic}[1]
        \renewcommand{\algorithmicrequire}{\textbf{Input:}}
        \renewcommand{\algorithmicensure}{\textbf{Output:}}
        \REQUIRE length of material $L$ (m) , coefficient of friction $k$, weight per unit length of material $q$ (kg)
        \ENSURE  Trajectory
        \FOR{$i$ in $\{0, 0.001, 0.002, \ldots, L\}$}
            \STATE $z_1 \gets i$
            \STATE Solve (\ref{deqn_ex9}) numerically for $\frac{l_1}{a}$
            \STATE Solve (\ref{deqn_ex7}) numerically for $a$
            \STATE $l_1 = a \cdot \frac{l_1}{a}$
            \STATE Solve (\ref{deqn_ex11}) numerically for $x_{1A}$
            \STATE Solve (\ref{deqn_ex10}) numerically for $\alpha_t$
            \STATE Calculate quaternion from Euler angles $\alpha_t$
            \STATE Save $x_{1A}$, $z_1$, quaternion of $\alpha_t$ as Waypoints
        \ENDFOR
        \RETURN Waypoints
        \STATE Generate Trajectory using ROS Moveit Motion Planning API with Waypoints
    \end{algorithmic}
\end{algorithm}

\subsection{Methodology of conducting experiments}
\noindent In automated lines handling textile and deformable objects, conveyor systems often utilize a special rubber covering (Fig.~\ref{fi1_1}) designed to generate a high coefficient of friction\footnote{Available: \url{https://romanmykhailyshyn.github.io/portfolio/portfolio-1/}}, preventing the objects from shifting during operation. Similarly, when deformable materials in a stack, friction between the layers plays a key role in maintaining stability. To replicate these real-world conditions, a textured neoprene covering with a crisscross pattern was selected, ensuring a high coefficient of friction between the deformable object and the surface.


\begin{figure}[!t]
    \centering
    \vspace{-3mm}
    \includegraphics[width=0.86\linewidth,clip ,trim=0pt 0pt 80pt 0pt]{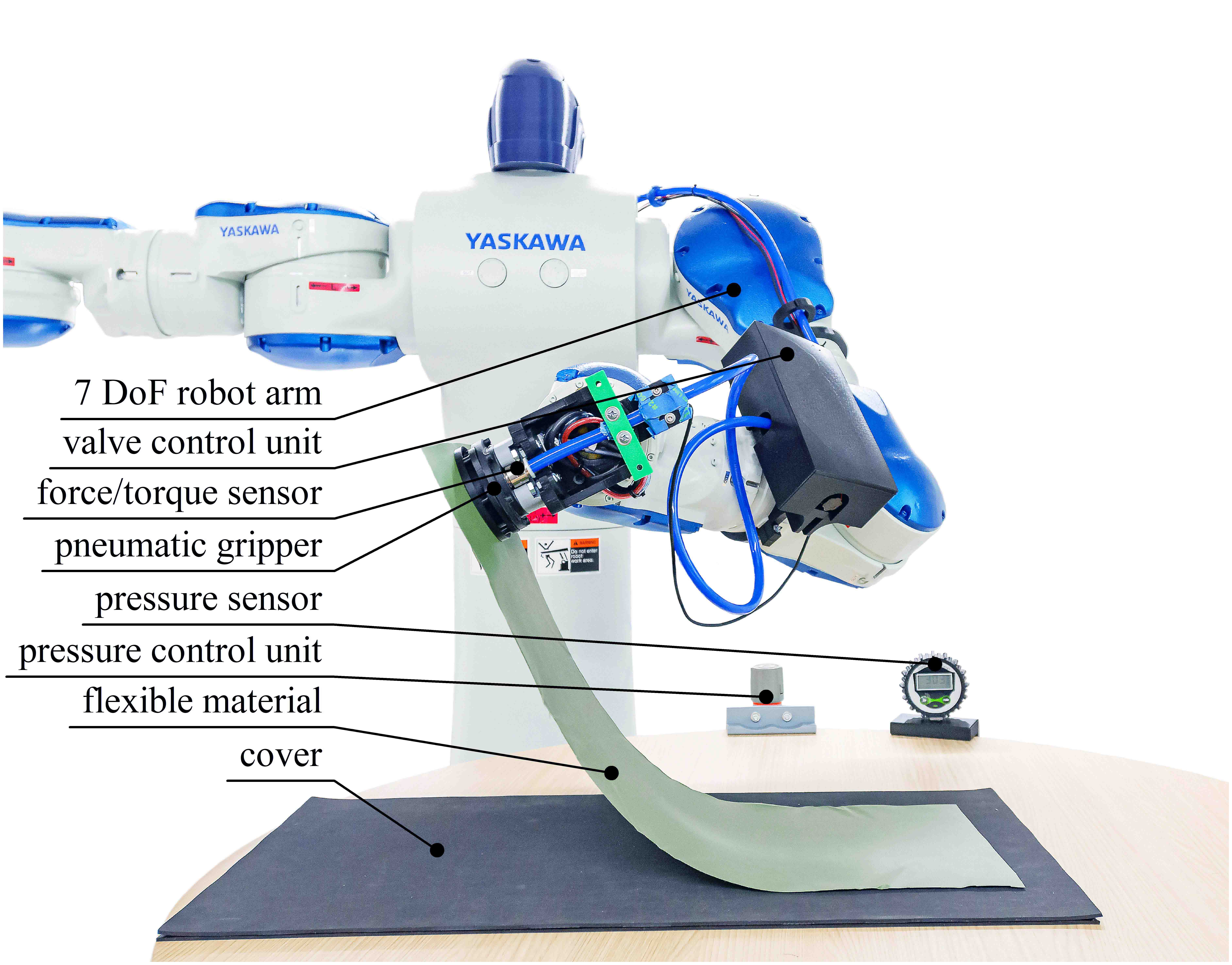}
    \caption{An experimental setup for the study of the effectiveness of manipulating deformable objects at one end using pneumatic grippers.}
    \label{fig_7}
\end{figure}

\begin{figure}[!t]
\newcommand{\mywidth}{1}
\centering
        \begin{subfigure}[b]{0.36\linewidth}
         \centering
         \vspace{1mm}
         \includegraphics[width=\mywidth\linewidth, clip, trim=0pt 0pt 0pt 0pt]{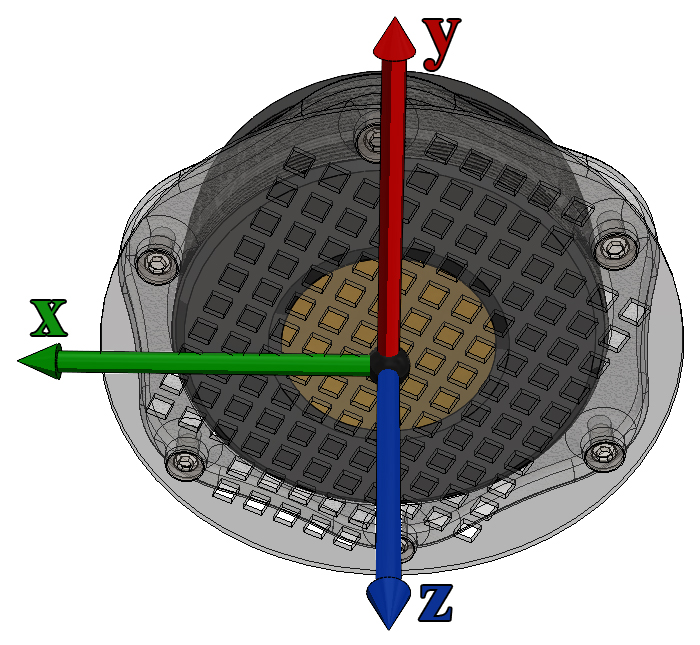}
         \vspace{-7mm}
         \caption{}
        \label{fig_8_3}
	\end{subfigure}	
        ~
        ~
        ~
        ~
        ~
  	\begin{subfigure}[b]{0.39\linewidth}
         \centering
         \vspace{1mm}
         \includegraphics[width=\mywidth\linewidth, clip, trim=0pt 0pt 0pt 0pt]{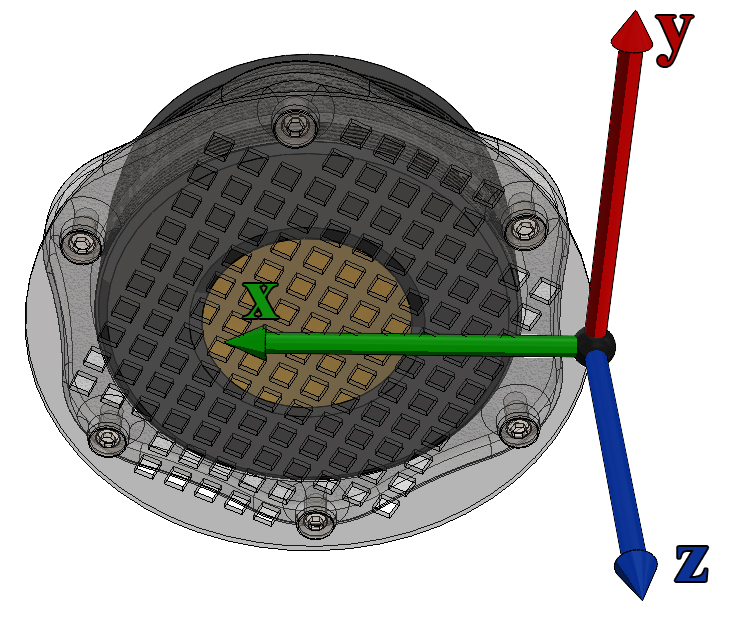}
         \vspace{-7mm}
        \caption{}
        \label{fig_8_2}
	\end{subfigure}
        \\
        \vspace{-1.5mm}
\caption{Position and orientation of the pneumatic gripper tool center point (TCP): (a) classical positioning on the central axis of the gripper; (b) for the proposed method, on the edge of the gripper.}
\label{fig_8_1}
\end{figure}

An experimental setup was developed to study the effectiveness of lifting deformable objects by one end using pneumatic grippers (Fig.~\ref{fig_7}). The experimental setup consists of a Yaskawa SDA10F robot to which a pneumatic gripper \cite{Mykhailyshyn2022gripper} is connected through an adapter with a force/torque ATI Nano-17 sensor with measurement limits for $F_x$ and $F_y$ 25 N for $F_z$ 35 N, boasting an accuracy of ±0.25$\%$. The deformable object lies in a predetermined position on the surface, without touching the object's edge, the pneumatic gripper is positioned at a distance of 5 mm from it. Then, according to Algorithm~\ref{alg1}, the force/torque sensor is calibrated and starts recording a new set of data, in parallel with this, a signal is sent to the valve control unit to supply compressed air (adjusted by the pressure control unit) to the gripper, which leads to the grasping of the deformable object. Continuing Algorithm~\ref{alg1}, the calculated position and orientation (trajectory) of the gripper's tool center point (TCP) are sent for execution to lift the deformable object. To implement the proposed methodology, the tool center point of the gripper is shifted to the edge of the gripper (Fig.~\ref{fig_8_2}) compared to the classical placement of the TPC in the center of the gripper (Fig.~\ref{fig_8_3}).

\begin{figure}[!t]
\newcommand{\mywidth}{1}
\centering
        \begin{subfigure}[b]{0.48\linewidth}
         \centering
         \vspace{-3mm}
         \includegraphics[width=\mywidth\linewidth, clip, trim=0pt 0pt 0pt 0pt]{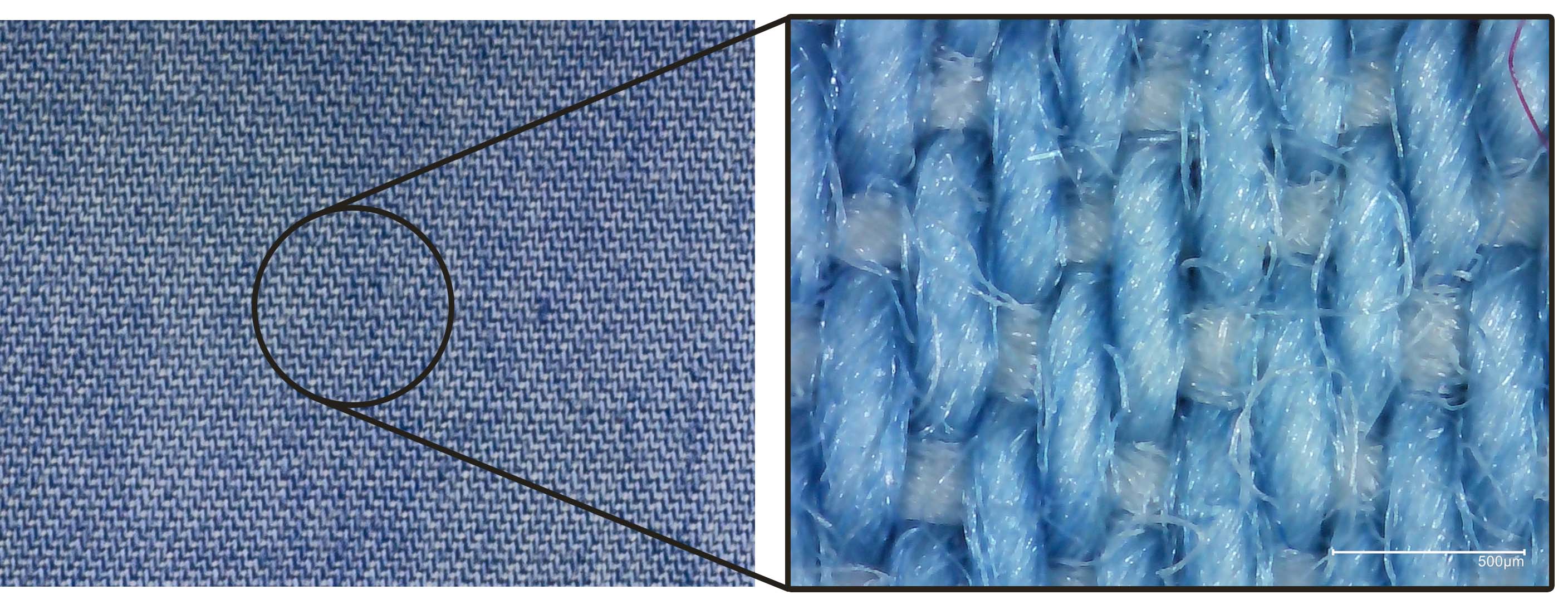}
         \vspace{-6mm}
         \caption{}
        \label{fig8_1}
	\end{subfigure}	
  	\begin{subfigure}[b]{0.48\linewidth}
         \centering
         \vspace{-3mm}
         \includegraphics[width=\mywidth\linewidth, clip, trim=0pt 0pt 0pt 0pt]{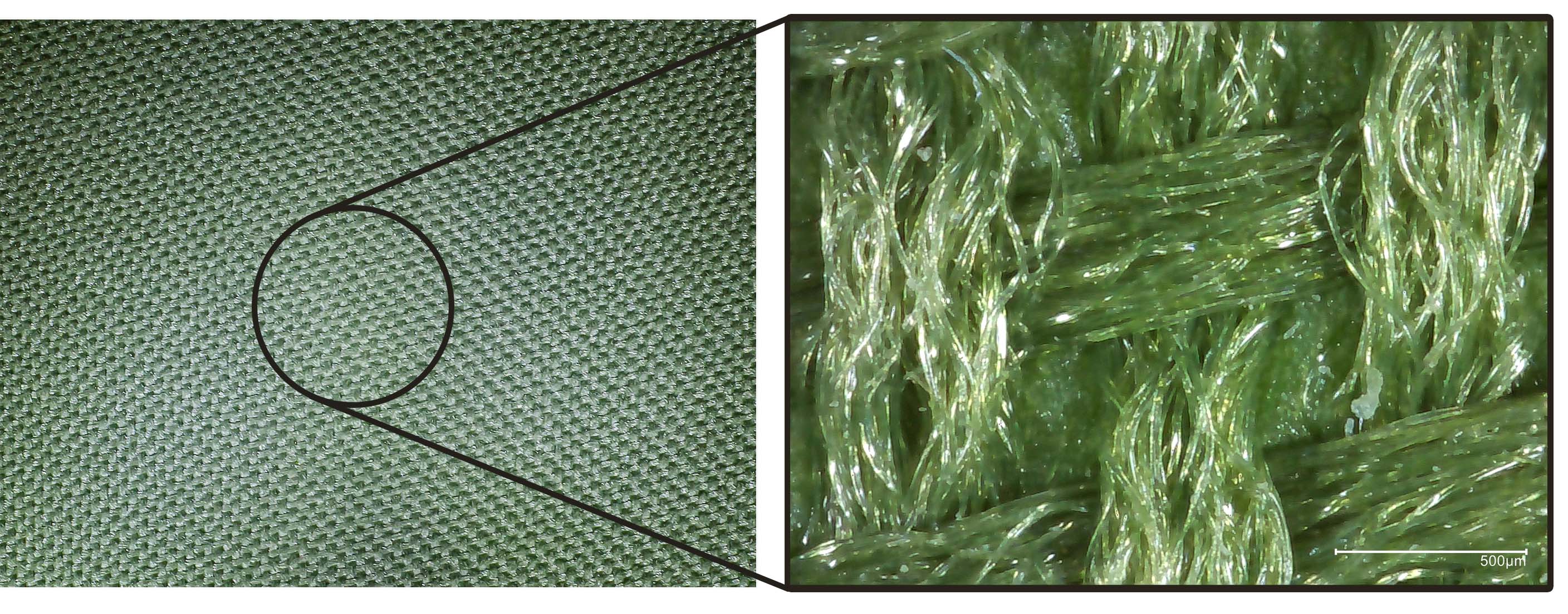}
         \vspace{-6mm}
        \caption{}
        \label{fig8_2}
	\end{subfigure}
        \\
        \begin{subfigure}[b]{0.48\linewidth}
         \centering
         \includegraphics[width=\mywidth\linewidth, clip, trim=0pt 0pt 0pt 0pt]{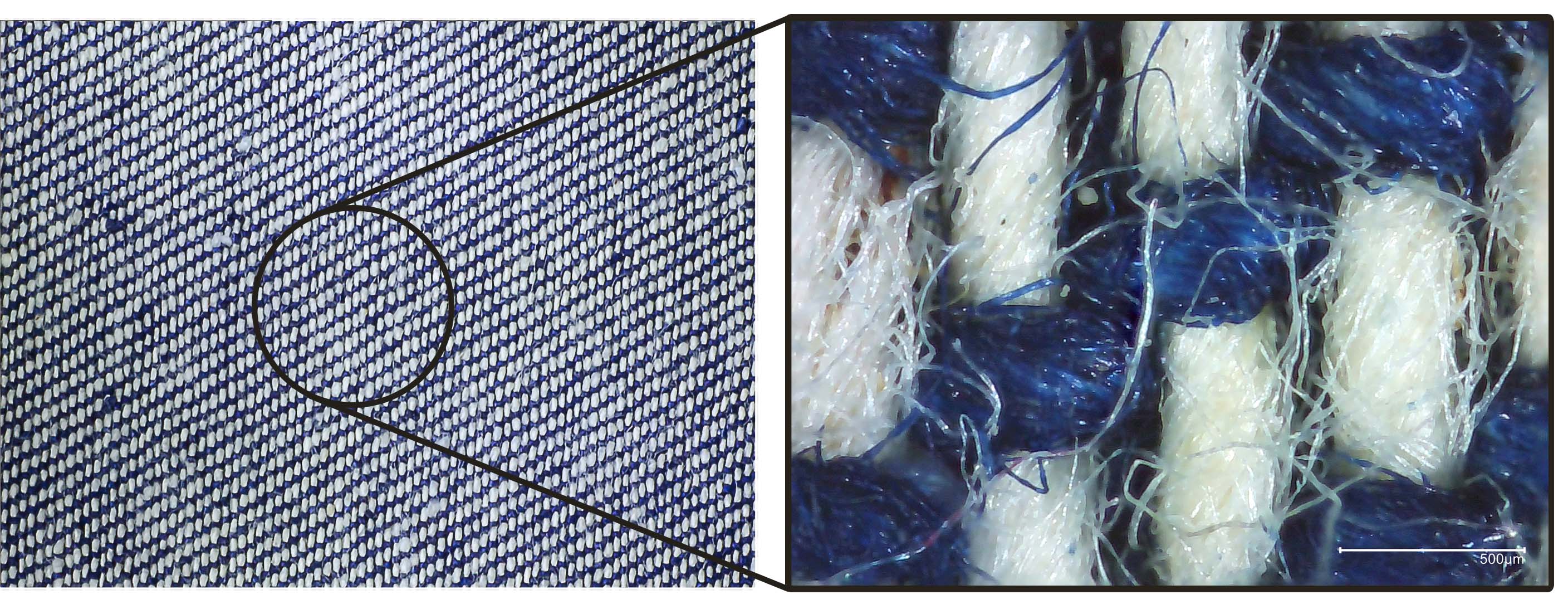}
        \vspace{-6mm}
         \caption{}
        \label{fig8_3}
	\end{subfigure}	
  	\begin{subfigure}[b]{0.48\linewidth}
         \centering
         \includegraphics[width=\mywidth\linewidth, clip, trim=0pt 0pt 0pt 0pt]{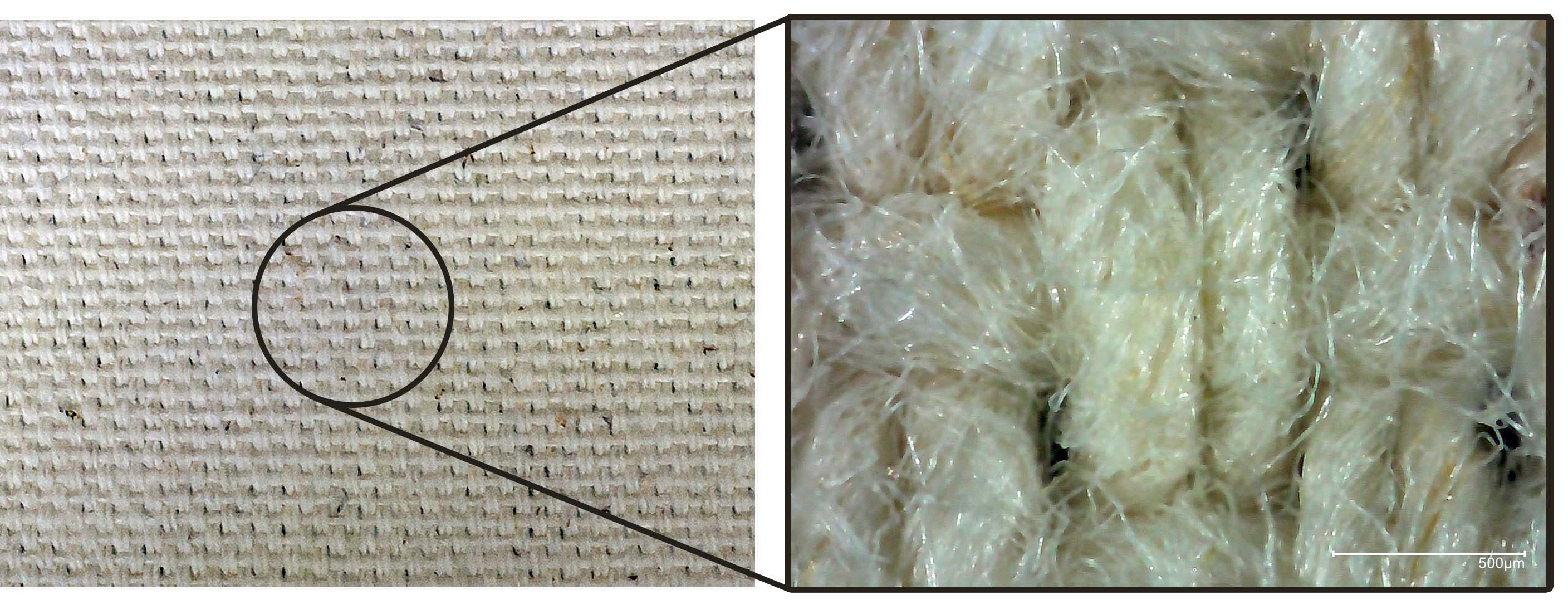}
        \vspace{-6mm}
        \caption{}
        \label{fig8_4}
	\end{subfigure}
        \\
        \vspace{-1.5mm}
\caption{Microscopy x20 and x150 of the studied textile materials: (a) \textnumero 1 thin cotton; (b) \textnumero 2 denier cordura; (c) \textnumero 3 heavyweight cotton; (d) \textnumero 4 heavyweight natural cotton.}
\label{fig_8}
\end{figure}

In order to ensure a comprehensive study of dexterous lifting deformable objects, four materials were selected (size 50x500~mm) for conducting the experiment with different properties (Table~\ref{tab1}): 
\begin{itemize}
\item №1 Thin Denim Cotton (Fig.~\ref{fig8_1}) - porous easily deformable light material.
\item №2 600-Denier Cordura Outdoor Canvas Waterproof Fabric (Fig.~\ref{fig8_2}) - non-porous hard-to-deform light material.
\item №3 Heavyweight Denim Cotton (Fig.~\ref{fig8_3}) - porous deformable heavy material.
\item №4 Heavyweight Denim Natural Cotton(Fig.~\ref{fig8_4}) - porous hard-to-deform heavy material.
\end{itemize}

\begin{table}
\begin{center}
\caption{Parameters of the Studied Materials.}
\label{tab1}
\begin{tabular}{| c | c | c | c | c |}
\hline
Parameters $\backslash$ Materials  & \textnumero 1 & \textnumero 2 & \textnumero 3 & \textnumero 4\\
\hline
$f$ with gripper & 0.50 & 0.57 & 0.35 & 0.44\\
\hline
Weight $m\cdot 10^-{}^4 $ [g/cm$^2$] & 0.014 & 0.031 & 0.036 & 0.031\\
\hline
Thickness of material $h$ [mm]  & 0.28 & 0.4 & 0.74 & 0.62\\
\hline
$k$ with covering & 1.54 & 1.71 & 1.49 & 1.38\\
\hline
\end{tabular}
\end{center}
\end{table}

\begin{figure}[!t]
    \centering
    \vspace{-2mm}
    \includegraphics[width=0.94\linewidth,clip ,trim=0pt 0pt 0pt 0pt]{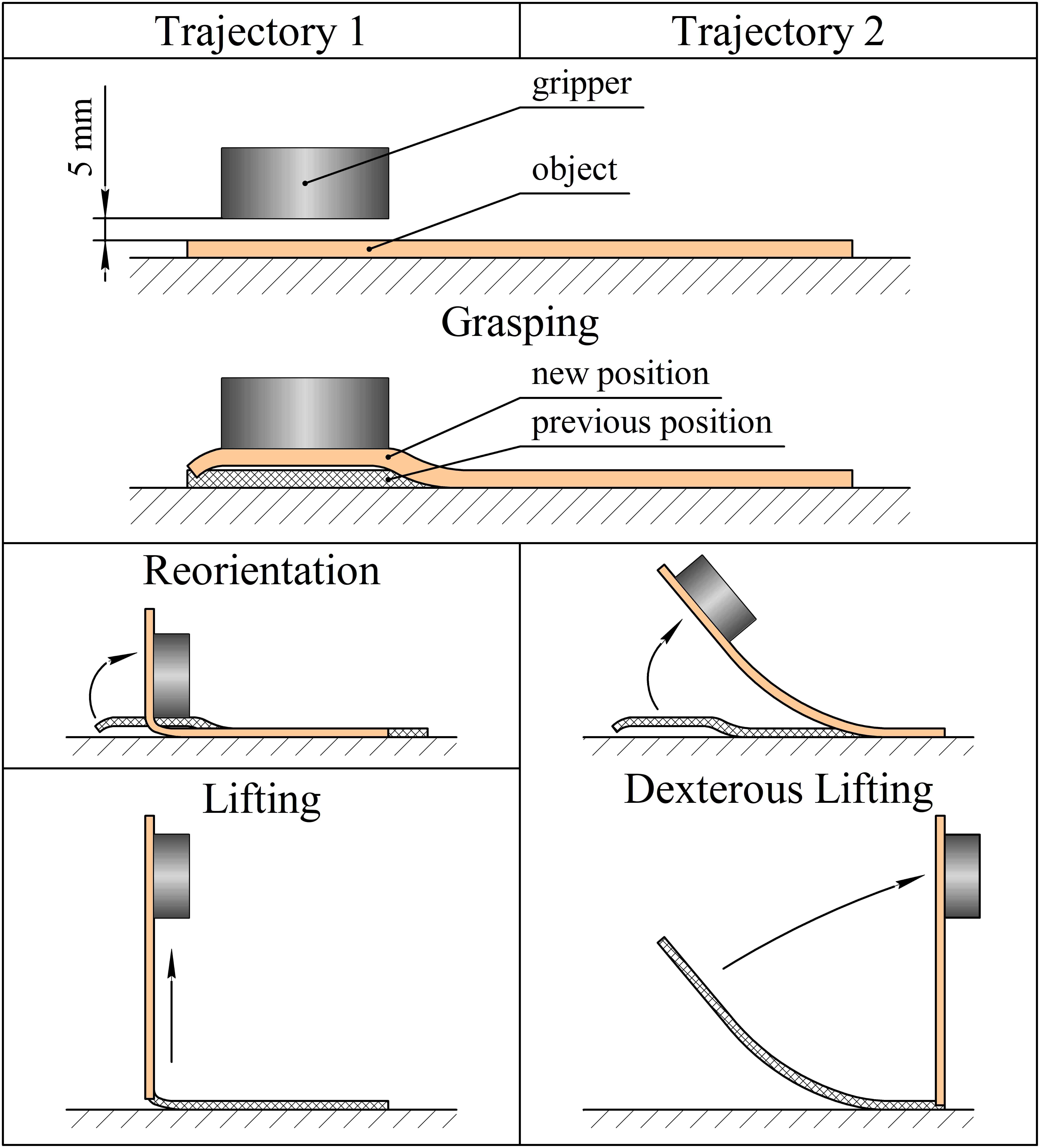}
    \caption{Diagram of the studied trajectories: Trajectory 1 ($T_1$) - reorientation and lifting; Trajectory 2 ($T_2$) - dexterous lifting.}
    \label{fig_9}
\end{figure}

An experimental study of the effectiveness of lifting deformable materials consisted of determining the minimum necessary supply pressure of the gripper for 100 percent success. Obviously, the gripper's minimum required supply pressure will always be greater than the pressure required to hold the object at the end point of the trajectory (at orientation $\alpha = 90$ degrees). Therefore, the minimum supply pressure~($P_0$) necessary to hold the deformable object~\cite{Mykhailyshyn2024toward} is first determined, and then an experiment is carried out by lifting the object, if it is not successful, the supply pressure is increased by 5 kPa ($P = P_{0} + 5$). This continues until the moment when we do not get a successful result, this result is checked 10 more times and if all the times of lifting the objects were successful, this pressure is fixed as the minimum necessary. If at least one of the 10 conducted experiments was not successful, the supply pressure is increased again by 5 kPa and the procedure is repeated.

\section{Result and Discussion}

\noindent During the classical lifting (Fig.~\ref{fig2}), all deformable materials falls. Therefore, we consider two lifting methods (Fig.~\ref{fig_9}):
\begin{itemize}
\item Trajectory 1 ($T_1$) with classical TCP (Fig.~\ref{fig_8_3}) - Grasping $\rightarrow$ Reorientation ($\alpha = 90$ degrees) $\rightarrow$ Vertical lift.
\item Trajectory 2 ($T_2$) with shifted TCP (Fig.~\ref{fig_8_2}) - Grasping $\rightarrow$ Dexterous Lifting (Algorithm~\ref{alg1}).
\end{itemize}

\begin{figure}[!t]
\newcommand{\mywidth}{1}
\centering
 	\begin{subfigure}[b]{0.96\linewidth}
         \centering
         \vspace{2mm}
        \includegraphics[width=\mywidth\linewidth, clip, trim=50pt 0pt 0pt 50pt]{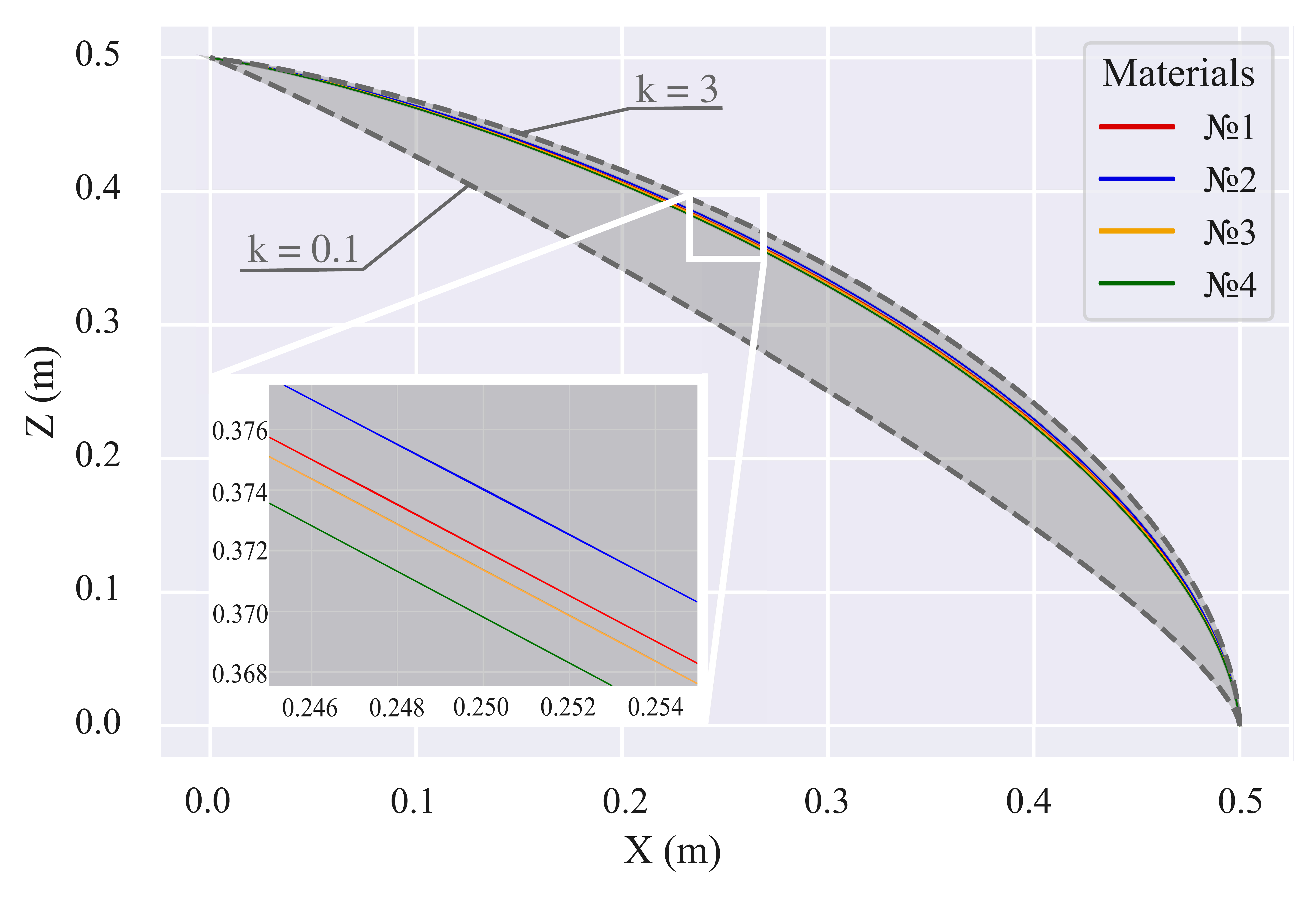}
        \vspace{-7.5mm}
        \caption{}
        \label{subfig10_1}
	\end{subfigure}\\
  	\begin{subfigure}[b]{0.96\linewidth}
         \centering
         \vspace{1mm}
         \includegraphics[width=\mywidth\linewidth, clip, trim=15pt 0pt 0pt 10pt]{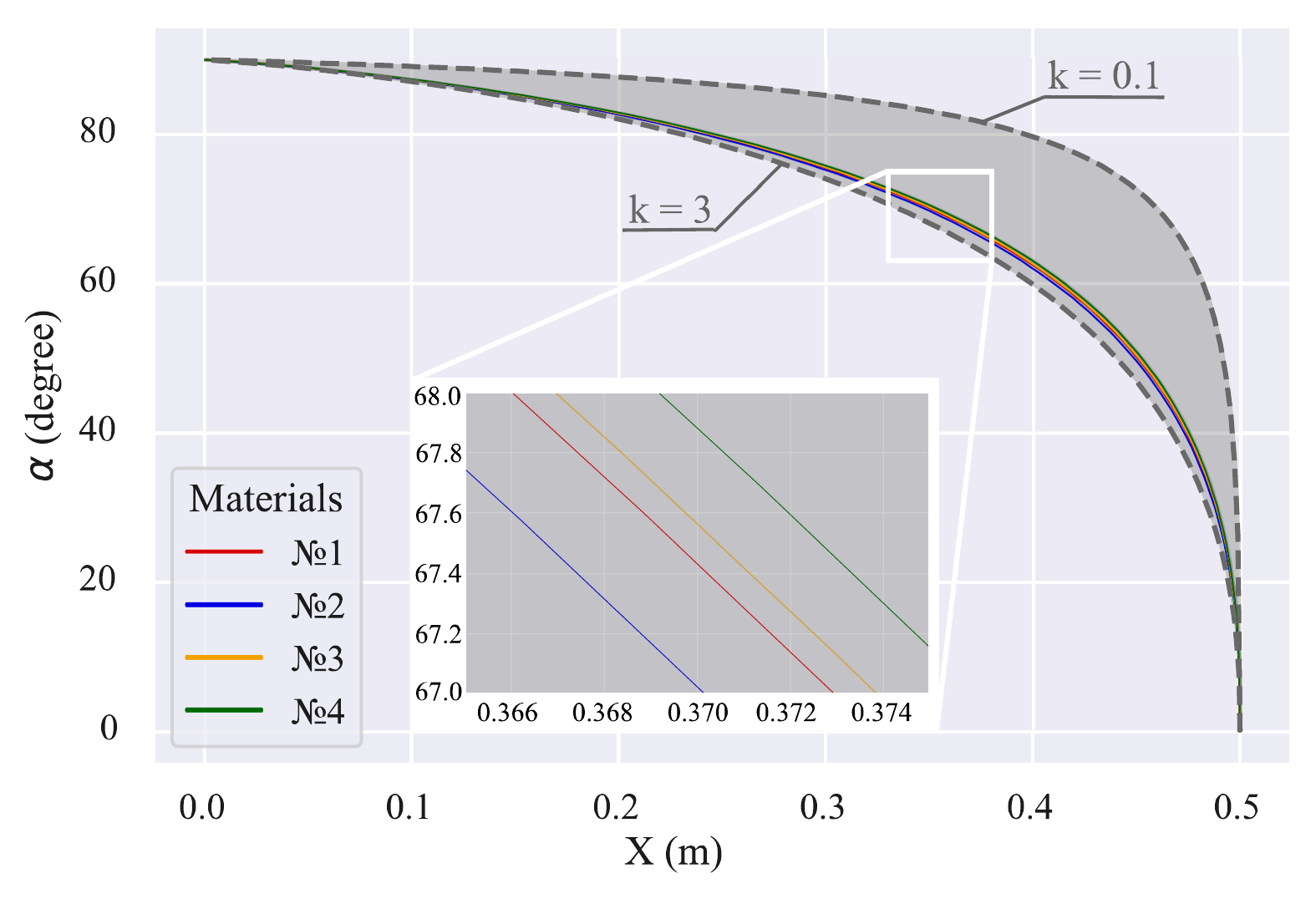}
         \vspace{-8mm}
        \caption{}
        \label{subfig10_2}
	\end{subfigure}\\
        \vspace{-2mm}
 \caption{Parameters of the dextrous lifting trajectory $T_2$ depending on the friction coefficient of the studied materials: (a) position along x and z axes; (b) orientation $\alpha$ relative to the x-axis.}
  \label{fig10}
\end{figure}

\begin{figure}[!t]
    \centering
    \vspace{2mm}
    \includegraphics[width=0.98\linewidth,clip ,trim=0pt 0pt 0pt 0pt]{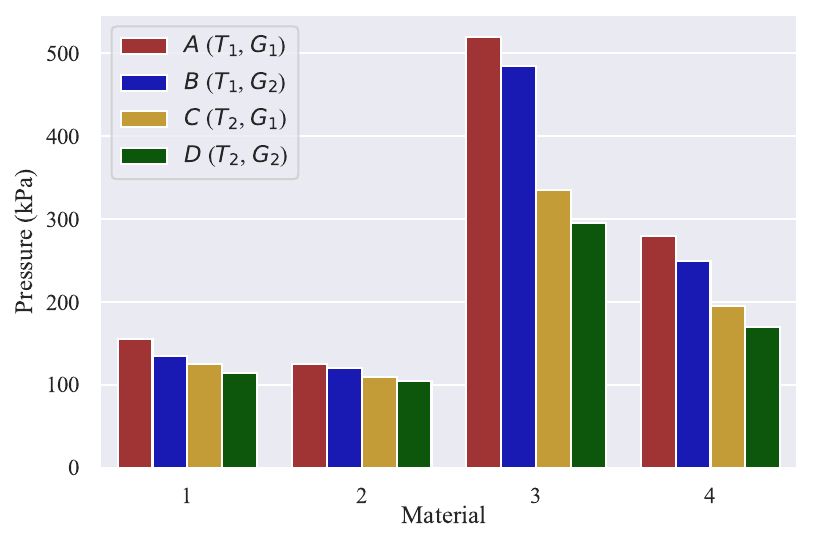}
    \vspace{-4mm}
    \caption{The minimum necessary gripper supply pressure that is necessary to hold the deformable object during different lifting methods of the studied materials.}
    \label{fig11}
\end{figure}

Also, to minimize vibration, we have improved the anti-vibration grid in the design of the gripper and we will now have two versions of the gripper ($G_1$ - with classic anti-vibration grid, $G_2$ - with rounded edge anti-vibration grid, to redirect air flows to the other side of the material), which allows us to conduct 4 options of the experiment for each material:
\begin{itemize}
\item A - Trajectory 1 ($T_1$) with Gripper 1 ($G_1$).
\item B - Trajectory 1 ($T_1$) with Gripper 2 ($G_2$).
\item C - Trajectory 2 ($T_2$) with Gripper 1 ($G_1$).
\item D - Trajectory 2 ($T_2$) with Gripper 2 ($G_2$).
\end{itemize}

Knowing the parameters, position, and dimensions of the materials (Table~\ref{tab1}) that we plan to lift, allowed us to use Algorithm~\ref{alg1} to find the parameters of the trajectory for each of the materials (Fig.~\ref{fig10}). It is obvious that one of the most important parameters affecting both the deflection of the deformable material and its position is the coefficient of friction between the material and the covering. Therefore, Fig.~\ref{subfig10_1} shows in the gray zone the change in the gripper's position during dextrose manipulation for a coefficient of friction ($k$) from 0.1 to 3. Accordingly, Fig.~\ref{subfig10_2} shows the gray zone for orientation of the gripper ($0.1 \leq k \leq 3$). 

Using the found trajectory (Fig.~\ref{fig10}) as $T_2$ for the studied materials according to the methodology described in Section II-B, we determined the minimum required gripper supply pressure for lifting each of the materials by all four methods (Fig.~\ref{fig11}).

\begin{figure}[!t]
    \centering
    \vspace{1mm}
    \includegraphics[width=0.84\linewidth,clip ,trim=0pt 0pt 0pt 0pt]{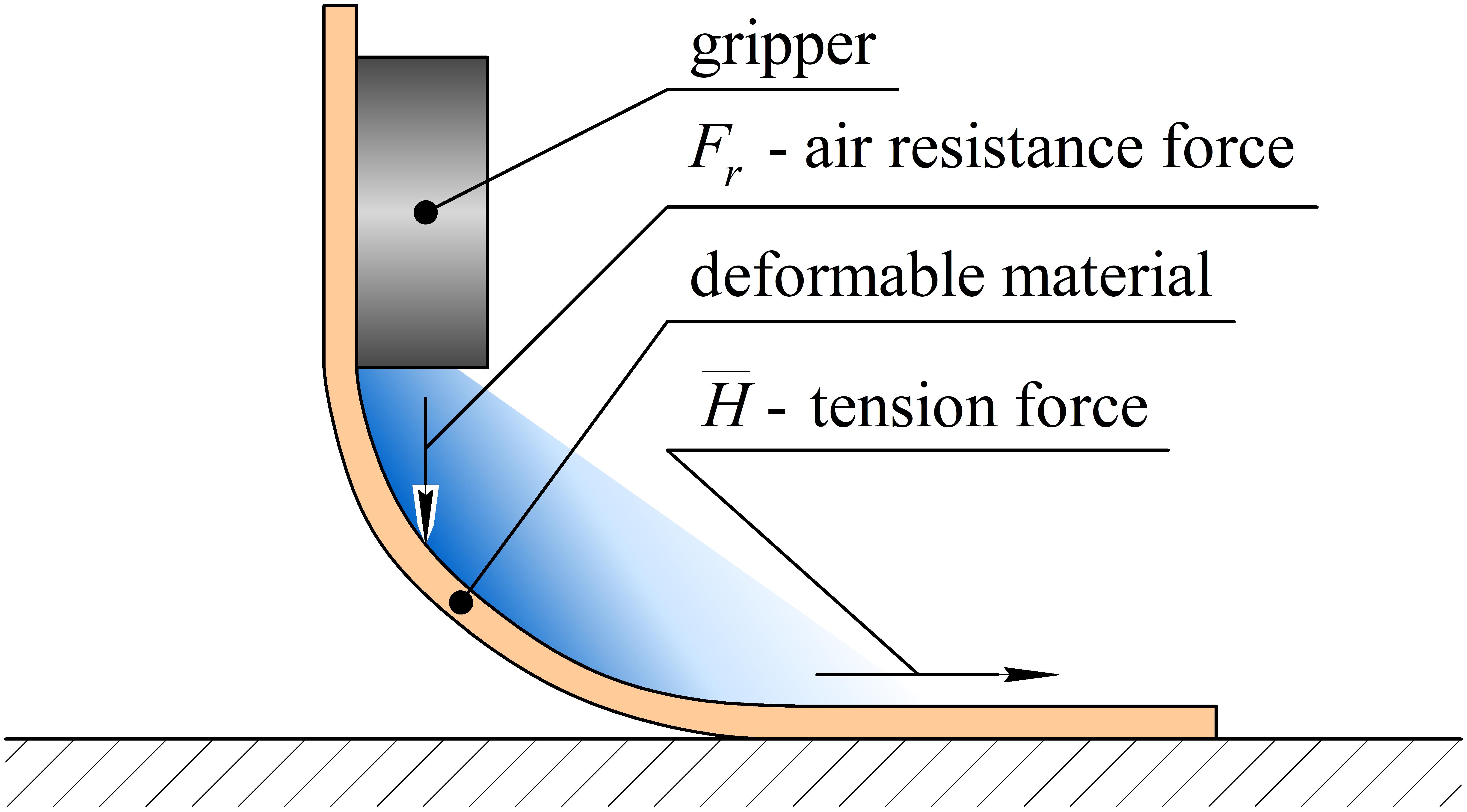}
    \vspace{-1.5mm}
    \caption{Additional forces arising when lifting deformable materials by the $T_1$ method.}
    \label{fig14}
\end{figure}

\begin{figure}[!t]
\newcommand{\mywidth}{1}
\centering
 	\begin{subfigure}[b]{1\linewidth}
         \centering
         \vspace{2mm}
        \includegraphics[width=\mywidth\linewidth, clip, trim=0pt 0pt 0pt 0pt]{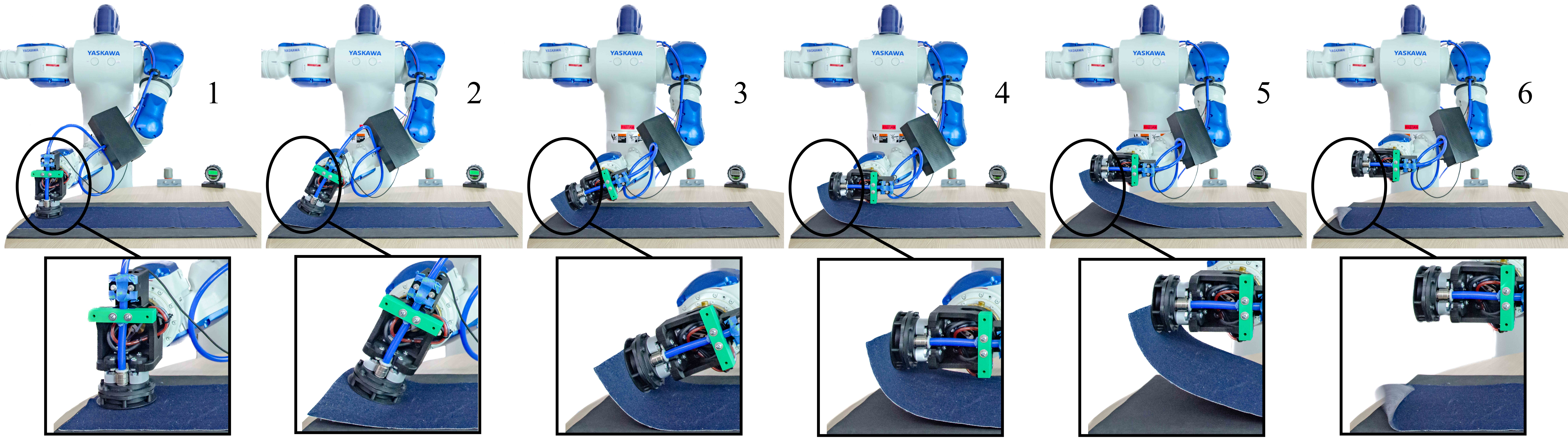}
        \vspace{-5mm}
        \caption{}
        \label{subfig13_1}
	\end{subfigure}\\
  	\begin{subfigure}[b]{1\linewidth}
         \centering
         \vspace{1mm}
         \includegraphics[width=\mywidth\linewidth, clip, trim=0pt 0pt 0pt 0pt]{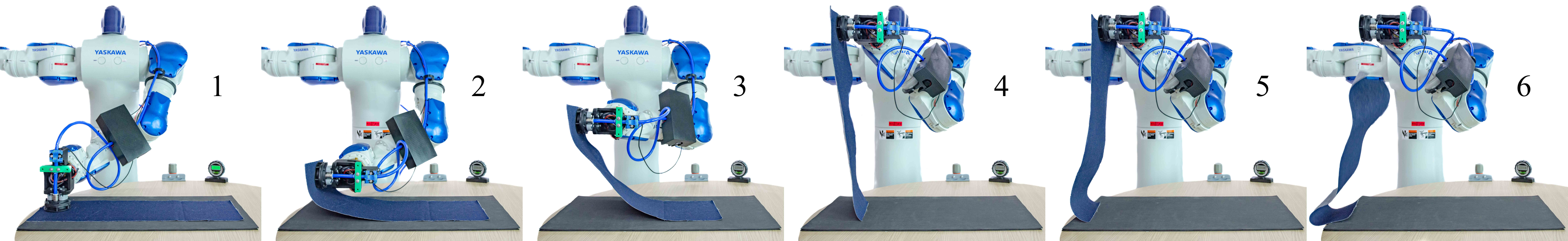}
         \vspace{-5mm}
        \caption{}
        \label{subfig13_2}
	\end{subfigure}\\
        \vspace{-1mm}
 \caption{Failing vertical lifting material \textnumero 3 using method A ($T_1$, $G_1$): (a) influence of the coefficient of friction between the covering and the textile on the sliding of the material along the gripper; (b) effect of the interaction of airflow from the gripper on the textile and its vibration.}
 \label{fig13}
\end{figure}

From Fig.~\ref{fig11}, it is obvious that the minimum required gripper supply pressure for lifting all materials by method A has the highest pressure values. This trend is most obvious for material \textnumero 3, where the minimum required gripper supply pressure for method A is 76\% higher than for method D. Also, for material \textnumero 1, the minimum required gripper supply pressure for method A is 35\% higher than for method D, for material \textnumero 2 by 19\%, and for material \textnumero 4 by 65\%. This has both numerical and analytical confirmations: 
\begin{itemize}
\item First of all, since the material is heavy and has a significant coefficient of friction (Table~\ref{tab1}), during the reorientation of the gripper ($T_1$), a significant tension force is formed (Fig.~\ref{fig14}), which leads to the sliding of the material along the gripper, loss of holding force and separation of the material from the gripper (Fig.~\ref{subfig13_1}).
\item Secondly, due to the design properties of the gripper ($G_1$), after reorientation of the gripper, the compressed air coming out of it collides with the deformable material leading to forming an air resistance force (Fig.~\ref{fig14}). As a result, it increases the minimum required lifting force needed to hold the material and creates unwanted vibration of the material (Fig.~\ref{subfig13_2}). 
\end{itemize}

\begin{figure}[!t]
    \centering
    \includegraphics[width=0.98\linewidth,clip ,trim=5pt 0pt 5pt 10pt]{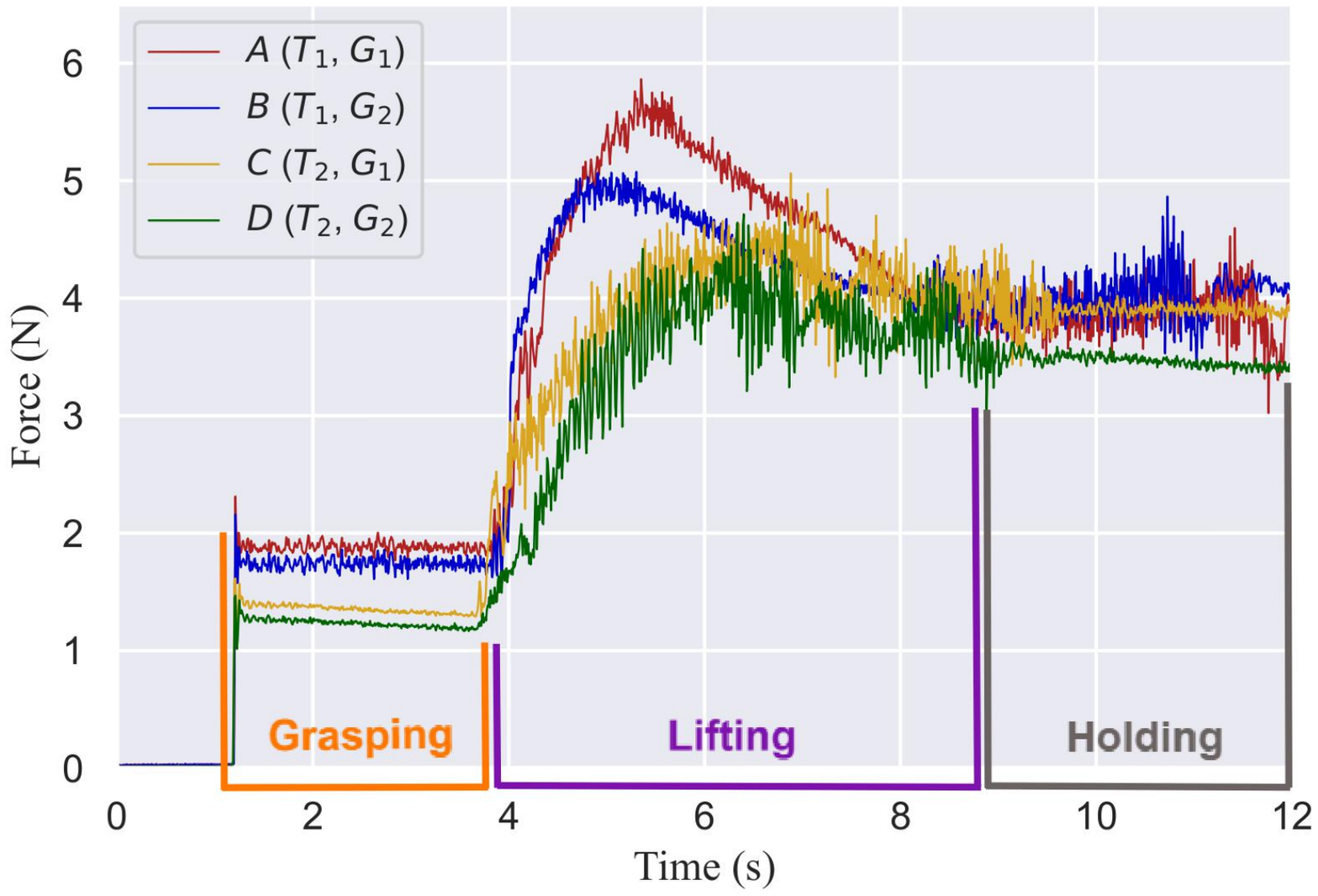}
    \vspace{-1mm}
    \caption{Change in the total force acting on the gripping device during manipulation of material \textnumero 4: Trajectory 1 ($T_1$) - reorientation and lifting; Trajectory 2 ($T_2$) - dexterous lifting; Gripper 1 $G_1$ - with classic anti-vibration grid, Gripper 2 $G_2$ - with beveled edge anti-vibration grid, to redirect air flows to the other side of the material.}
    \label{fig15}
\end{figure}

Analyzing the obtained minimum supply pressure necessary for lifting, we see the trend $P_A>P_B>P_C>P_D$ for all the studied materials. In addition, it is obvious from the results that the influence of the trajectory on the minimum necessary supply pressure for lifting is the most significant. For a more detailed analysis, we analyzed the total vector of forces acting on the gripper during the lifting of material \textnumero 4 by various methods (Fig.~\ref{fig15}).

Analyzing the total force acting on the gripping device (Fig.~\ref{fig15}), it is possible to clearly distinguish the stages of grasping, lifting/manipulating, and holding. The gripping stage is practically identical for all lifting methods and differs only in a slight decrease in force for methods C and D. This is explained by the fact that for the two methods C and D, manipulation can be carried out with a lower supply pressure, which reduces the distance from which the deformable material is gripped. Since the gripping distance decreases for the C and D method, the amount of material that is gripped decreases accordingly, which we see at the gripping stage leads to a decrease in the total force.

The stage of lifting/manipulating a deformable object (Fig.~\ref{fig15}) allows us to evaluate the influence of the trajectory and the gripper on the effectiveness of this operation. It is obvious that there is an increase in force during the reorientation stage of method A, which leads to an increase in the required holding force. Since at the stage of lifting/manipulation, the force is maximum, this is the critical moment when the deformable material can separate from the gripper and fall. The reduction in force between the A and B methods is the largest since the air front resistance force acting on the deformable object in method A plays a significant role. Accordingly, each subsequent method B, C, and D has a smoother transition to holding. By reducing the force jump during the lift/manipulation stage, the required gripper supply pressure is managed to be reduced to perform this operation. 

The holding stage (Fig.~\ref{fig15}) shows the residual vibration of the deformable material after lifting. It is obvious that the choice of the trajectory has the greatest influence, which for methods A and B ($T_1$) is obvious because we see force jumps. On the other hand, residual vibration is practically absent for the C and D methods ($T_2$), which in general leads to a decrease in forces. Also, let's compare methods A, and C with B, and D. We can see a decrease in force/vibration jumps in B and D. Using methods A and C can lead to material falling due to the minimization of the coefficient of friction between the deformable material and the gripper (Fig.~\ref{subfig13_2}). Therefore, choosing a gripping device ($G_2$) with a redirection of the airflow from the grasping material is obvious.

\begin{figure}[tb]
    \centering
    \vspace{0.5mm}
    \includegraphics[width=1\linewidth,clip ,trim=0pt 0pt 0pt 0pt]{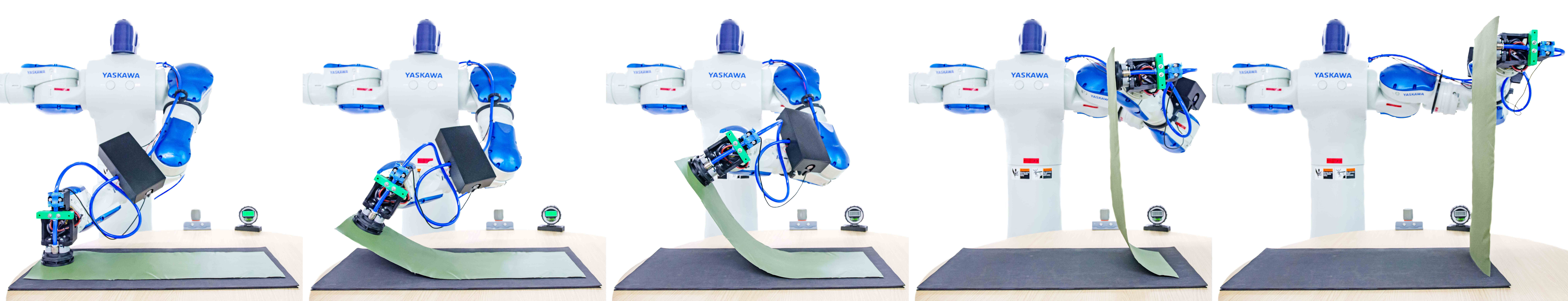}
    \caption{Successful dexterous lifting of deformable material \textnumero 2 at a minimum required gripper supply pressure of 105 kPa.}
    \label{fig_16}
\end{figure}

The obtained results of the experiment on the minimum necessary supply pressure (Fig.~\ref{fig11}) and force characteristics of the lifting/manipulation process (Fig.~\ref{fig15}) allow us to state that method D is the most effective. Method D aims to use the proposed dextrose lifting/manipulation of deformable objects and the gripper with a redirected direction of airflow from the material. This method allows you to avoid the formation of additional forces (air resistance, tension), which cause an increase in energy costs for the lifting process, and to carry out previously unavailable dexterous manipulation of deformable materials in manufacturing (Fig.~\ref{fig_16}).

\section{Conclusion}
This paper introduces a novel method for dexterous lifting of deformable materials using pneumatic grippers. Unlike previous methods, this approach involves adjusting the position and orientation of the gripper during the lifting process. This technique effectively minimizes the interaction of airflows with the deformable object, preventing detachment from the gripper. Additionally, the improved gripper design, which redirects airflow away from the material, significantly reduces vibrations during handling. Overall, the implementation of this method enables the dexterous lifting of deformable objects with minimal energy consumption, which was previously unavailable with pneumatic grippers\footnote{[Online]. Available: \url{https://youtu.be/ZD-lK-SoFko}}. Depending on the material type, the required gripper supply pressure was reduced by 19\% to 76\% per lift. This method has the potential to enhance automation and lower costs in complex manufacturing processes involving deformable objects.


Future work will focus on extending this technique to handle more complex object shapes and integrating artificial vision systems for greater adaptability in manufacturing. Additionally, efforts will be made to optimize the gripper's anti-vibration grid parameters to further improve airflow redirection and increase lifting force.


%



\bibliographystyle{IEEEtran}
\IEEEtriggercmd{\enlargethispage{2in}}
\bibliography{Dex}

\end{document}